\documentclass[journal]{IEEEtran}

\usepackage{amsmath,amsfonts}
\usepackage{algorithmic}
\usepackage{algorithm}
\usepackage{array}
\usepackage[caption=false,font=normalsize,labelfont=sf,textfont=sf]{subfig}
\usepackage{textcomp}
\usepackage{stfloats}
\usepackage{url}
\usepackage{verbatim}
\usepackage{graphicx}
\usepackage{cite}
\usepackage{caption} 

\usepackage{booktabs}
\usepackage{adjustbox}
\usepackage{multirow}
\usepackage{ulem}
\usepackage[para,online,flushleft]{threeparttable}
\usepackage[colorlinks,linkcolor=blue]{hyperref}
\usepackage{setspace}
\usepackage{xcolor,soul,framed} 
\colorlet{shadecolor}{yellow}
\usepackage{array}
\usepackage{mdwmath}
\usepackage{mdwtab}
\usepackage{eqparbox}
\usepackage{url}
\usepackage{multirow}
\usepackage{authblk}
\usepackage{adjustbox}
\usepackage{multirow}
\usepackage{ulem}
\usepackage{fancyhdr} 
\usepackage{setspace}

\newcommand{\etal}{\textit{et al.}}

\begin{document}
\title{Data Hiding with Deep Learning: A Survey Unifying Digital Watermarking and Steganography}

\author[1,4]{Zihan Wang}
\author[2]{Olivia Byrnes}
\author[2]{Hu Wang}
\author[1]{Ruoxi Sun}
\author[2]{Congbo Ma}
\author[3]{Huaming Chen}
\author[2]{Qi Wu}
\author[1]{Minhui Xue}
\affil[1]{CSIRO's Data61, Australia}
\affil[2]{The University of Adelaide, Australia}
\affil[3]{The University of Sydney, Australia}
\affil[4]{The University of Queensland, Australia}

\maketitle

\lfoot{}
\renewcommand{\headrulewidth}{0mm}

\begin{abstract}
The advancement of secure communication and identity verification fields has significantly increased through the use of deep learning techniques for data hiding. By embedding information into a noise-tolerant signal such as audio, video, or images, digital watermarking and steganography techniques can be used to protect sensitive intellectual property and enable confidential communication, ensuring that the information embedded is only accessible to authorized parties. This survey provides an overview of recent developments in deep learning techniques deployed for data hiding, categorized systematically according to model architectures and noise injection methods. The objective functions, evaluation metrics, and datasets used for training these data hiding models are comprehensively summarised. Additionally, potential future research directions that unite digital watermarking and steganography on software engineering to enhance security and mitigate risks are suggested and deliberated. This contribution furthers the creation of a more trustworthy digital world and advances Responsible AI.

 \end{abstract}


\IEEEpeerreviewmaketitle

\section{Introduction}
\label{section:1}

\IEEEPARstart{S}{erious} concerns are raised about the ability of Artificial Intelligence to make responsible decisions or behave responsibly, such as generating unfair outcomes, causing job displacement or insufficient protection of privacy and data security. In response, Responsible AI aims to address these issues and create accountability for AI systems.

Data hiding is considered a promising method to achieve data security towards Responsible AI. Typically, data hiding involves concealing information in a specific form within another type of media. It can take different forms, such as encoding confidential information into an existing text piece or embedding audio files into a digital image. As digital assets become more diverse and ubiquitous, the importance and scope of data hiding applications will only continue to grow~\cite{survey}.
In today's digital age, as digital communication and multimedia data become increasingly prevalent, the data hiding process has become crucial. Ensuring responsible AI, such as machine learning as services, per requisite, necessitates secure communication across all mediums, and the accountability of responsible AI, such as digital intellectual property, necessitates protection against theft and misuse. In its traditional form, the data hiding process can be categorized into three types: watermarking, steganography, and cryptography~\cite{dhsurvey}. This survey focuses on the former two (watermarking and steganography), as both have demonstrated superior capability in safeguarding sensitive data for AI systems learning.

Digital watermarking utilizes data hiding techniques to embed an identification (ID) into digital media that communicates with the owners of the intellectual property (IP) to prevent unauthorized copying or alteration. Thus, in case of any attempt to copy or modify the original media, the ID can be extracted to identify the owners. The primary application of digital watermarking is for the authentication of digital assets, however, the process has also been used for licences and identification \cite{licence}, digital forensics \cite{forensics}, and data protection in smart cities \cite{smartcity,iot}. Digital watermarking is not only useful for marking images and documents, but also for real-time audios and videos \cite{audiosurvey,videosurvey}, languages \cite{textwm}, as well as chip and hardware protection in electronics \cite{hardware}.

Steganography and watermarking share similarities in that both involve embedding data into a piece of media. However, while watermarking aims to identify the creator of an artifact, steganography embeds secret messages in a way that avoids detection, interception, or decoding. Unlike cryptography, which is designed to secure data by taking advantage of complexity, steganography's primary goal is to keep the cover media's format readable and not distorted after data hiding. The public should still be able to see the original cover media without noticing the embedded messages. Steganography is applied in various industries, including medicine, defense, and multimedia fields, wherever confidentiality is crucial for secure communication \cite{s4}.

\begin{figure*}[t]
    \centering
    \includegraphics[width=0.82\textwidth]{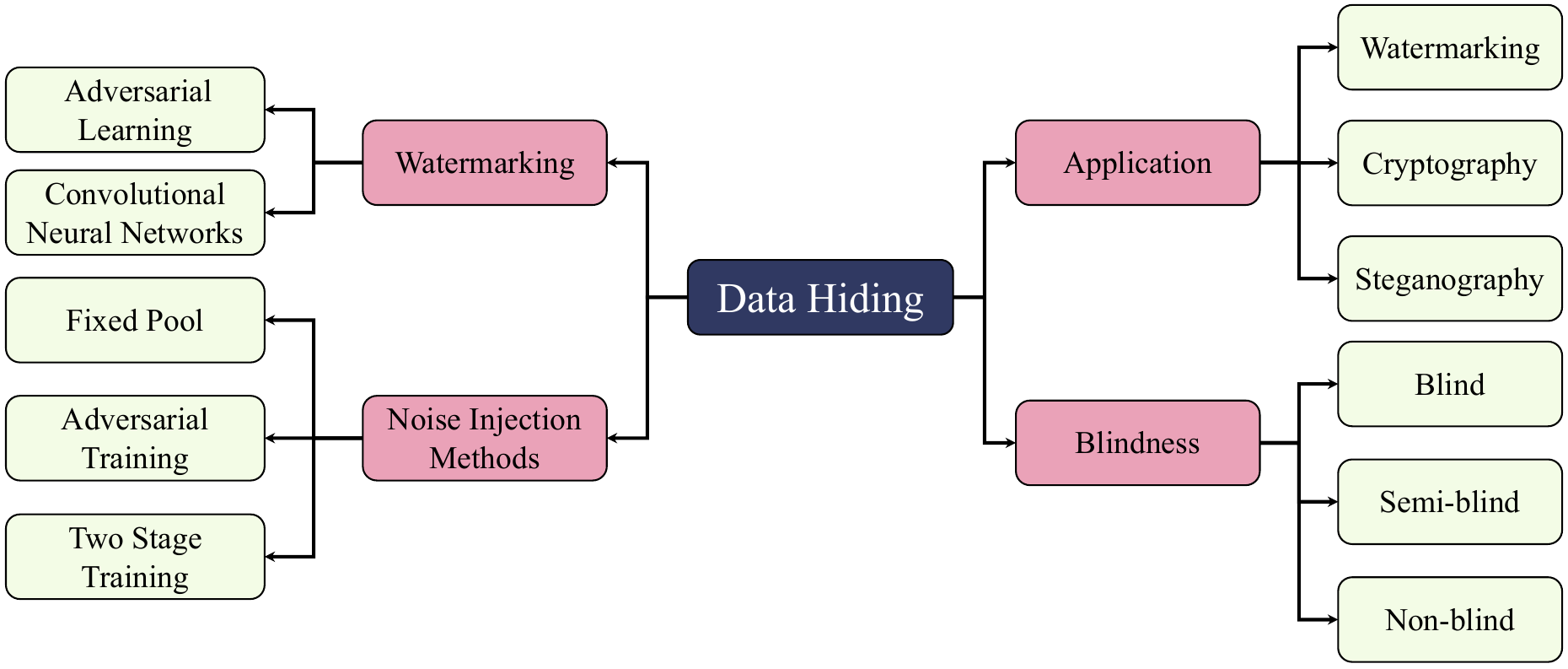}
    \caption{A hierarchical diagram showing different methods for classifying deep learning-based data hiding techniques. Blindness refers to the functionality of the data hiding method, further explained in Section \ref{section:2}. 
    }
    \label{fig:section1}
\end{figure*}

Historically, data hiding has been accomplished using specialized algorithms that are classified based on the domain in which they operate, either spatial or frequency. Spatial domain techniques directly embed data into the cover media by manipulating bit streams or pixel values. They are computationally simple compared to other techniques, and hence are more susceptible to removal and distortion from adversaries. Frequency domain techniques rely on the manipulation of frequency coefficients in the signal medium. These techniques achieve a higher degree of robustness to attacks, but are more computationally complex \cite{survey}. 

The drawback of these traditional algorithms is that their applications are narrow, and the creators of these algorithms require expert knowledge of the embedding process. Particular techniques are useful for certain limited tasks, and the growing sophistication of watermark removal and degradation attacks means that the effectiveness of these algorithms may be compromised in the near future \cite{firstCNN,wmremoval}. 

New advancements in the field of deep learning span many industries due to the strong representation abilities of deep neural networks. In the field of data hiding, deep learning models provide adaptable, generalized frameworks that can be used for a variety of applications in watermarking and steganography. Currently, most works in this area concentrate on image-based data hiding, and these are the works that will be compared in this survey. These machine learning models are able to learn advanced embedding patterns that are able to resist a much wider range of attacks with far more effectiveness than traditional watermarking or steganography algorithms \cite{firstCNN}. Further research on \cite{dvmark,mesh,textwm,hidden} established the potential and criticality of developing generalized frameworks for data hiding that can work with a range of cover media types to robustly embed data and provide highly secure content authentication and communication services. The advantage of the deep learning approach is that networks can be retrained to become resistant to new types of attacks, or to emphasise particular goals such as payload capacity or imperceptibility without creating specialised algorithms for each new application \cite{hidden}. An additional advantage of deep data hiding techniques is that they can enhance the security of the embedded messages. The high non-linearity of deep neural models makes it virtually impossible for an adversary to retrieve the embedded information~\cite{firstCNN}. 
Compared to traditional methods, deep learning-based methods are not only more secure and adaptable to different applications, but they also offer enhanced robustness to adversarial attacks and distortions. They are also able to achieve more imperceptible forms of data embedding.

The process of data hiding, comprising message embedding and extraction, can be naturally mapped onto an encoder-decoder network architecture. This approach involves partitioning the learning model into two networks: the encoder network, which learns to embed input messages into images, and the decoder network, which extracts the original message from the distorted image after subjecting it to various forms of attacks, such as blurring, cropping, compression, etc. The network is trained by minimizing an objective function that accounts for the differences between the cover image (i.e., the data carrier) and the encoded image, as well as the differences between the embedded and extracted input message.

The first papers exploring the capabilities of neural network technology for data hiding were released in 2017, and were based on convolutional neural networks \cite{firstCNN, hidingplain}. In recent years, GAN-based approaches have gained traction, popularised by the HiDDeN model \cite{hidden}, which was the first end-to-end trainable model for digital watermarking and steganography. These deep learning models employ different message embedding strategies in order to improve robustness, such as using adversarial examples, attention masks, and channel coding. The continued development of deep learning-based data hiding models will greatly improve the effectiveness and security of digital IP protection, and secure secret communication. 

Since this is a relatively new area of research, current surveys on data hiding primarily concentrate on traditional algorithms.
There are existing works examining deep learning-based techniques for steganography and cryptography \cite{cryptsurvey,stegsurvey}, but there is a lack of works examining deep watermarking techniques. 
There is an existing survey looking at deep learning-based watermarking and steganography~\cite{dlwmsteg}; however, a comprehensive survey regarding deep data hiding models unifying digital watermarking and steganography is still lacking. To the best of our knowledge, ours is the first survey to examine deep learning techniques for deep learning-based digital watermarking and steganography that includes the most extensive range of recent works. 
As this research area continues to expand, it is important to summarise and review the current methods. The survey aims to systematically categorise and discuss existing deep learning models for data hiding, separated based on applications in either watermarking or steganography, as well as present future directions that research may take. The key contributions of the survey are listed as follows:
\begin{itemize}
    \item This survey systematically categorises and compares deep learning-based models for data hiding in either watermarking or steganography based on network architecture and noise injection methods.
    \item We comprehensively discuss and compare different objective strategies, evaluation metrics, and training datasets used in current state-of-the-art deep data hiding techniques. 
    \item We also outline a broad spectrum of potential research avenues for the future development of deep learning-based data hiding.
\end{itemize}

\noindent \textbf{Paper Collation.} Our survey involved the collection, analysis, and discussion of over 30 papers retrieved from Google Scholar\footnote{https://scholar.google.com/} and DBLP\footnote{https://dblp.org/}, using the search keywords \textit{data hiding, digital watermarking, steganography, deep neural networks, and generative adversarial networks (GANs)}. Those papers were selected from a plethora of top-tier Security and Privacy, Computer Vision and Machine Learning conferences and journals. 

\noindent \textbf{Organization of the Survey.} In the following sections, our survey will cover recent advanced deep learning based data hiding methods in two forms: digital watermarking and steganography. Section \ref{section:2} gives the problem formulation of data hiding. Section \ref{section:3} highlights the architecture of data hiding and offers a comprehensive review of deep learning based data hiding techniques. This survey also summarises noise injection techniques in Section \ref{section:4}, objective functions in Section \ref{section:5}, evaluation metrics in Section \ref{section:6}, and existing datasets in Section \ref{section:7}. Finally, open questions and future work for deep learning based data hiding tasks are discussed in Section \ref{section:8}. Section \ref{section:9} concludes the paper.

\section{Problem Statement}

\label{section:2}

When we evaluate the effectiveness of data hiding techniques, there are many factors that should be considered. The three most important are 
\textbf{capacity}, how much information can be embedded into the cover media, \textbf{imperceptibility}, how easy the data is to detect, and \textbf{robustness}, how resistant the data is to attacks. Here, attacks refer to any alterations made to the embedded media with the intent to degrade or remove the embedded data. There is an implicit trade-off between these three aforementioned characteristics. For instance, if there is a high payload capacity, then the message will be easier to detect, resulting in a lower level of imperceptibility. Similarly, improving robustness against attacks can potentially decrease both payload capacity and imperceptibility, since there is added redundancy to the encoded image that allows it to resist distortions.
 
\begin{figure}  
\centering
\includegraphics[width=3.0in]{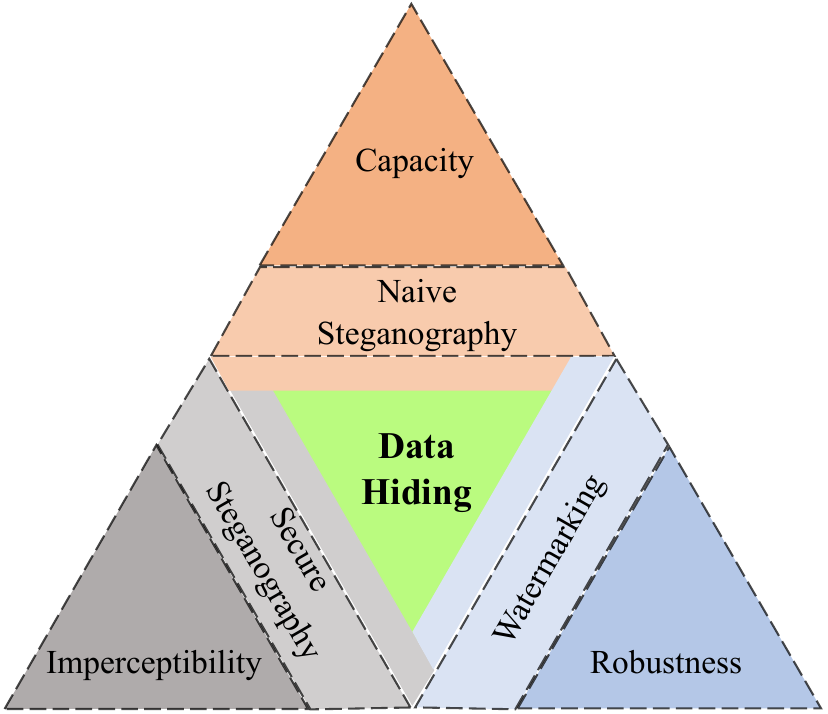}\\
\caption{A figure showing the trade-off between the three primary data hiding properties; robustness, imperceptibility, and capacity, as well as which data hiding applications favour each property over the others.}\label{fig:diagram2}
\end{figure}

In digital watermarking, robustness is generally favoured over secrecy because the ability to resist attacks and distortions is more important than the watermark's imperceptibility. Conversely, in steganography, imperceptibility is favoured since the highest priority is that the message remains a secret. This relationship is illustrated in Figure~\ref{fig:diagram2}. 
Due to the adaptable nature of deep learning-based approaches, the trade-off between these metrics can be explicitly controlled by the user, and the key properties of robustness and imperceptibility underpin the objective of the deep learning system. 
 
\subsection {Digital Watermarking and Steganography Terminology}

\label{section:2.1}

The basic data hiding process consists of an encoding and decoding process. The encoder $E$ receives the cover media $C$ and message to be hidden $M$, and outputs the encoded media $C'$ such that: $E(C,M) = C'$. 
Then the decoder receives the encoded media as input and extracts the message $M'$, such that: $D(C')=M'$. In a robust implementation, $M$ and $M'$ should be as similar as possible in an effective strategy. Similarly, maximising the imperceptibility property is done by minimising the difference between $C$ and $C'$. Types of data hiding can be classified based on a variety of properties as follows~\cite{survey}:

\begin{itemize}
  \item \textbf{Blindness:} related to how much information is required to extract the original data from the encoded media. \textbf{Blind} techniques do not require the original cover media or original data to extract the embedded data, and are the most practically useful. \textbf{Semi-blind} techniques require only the original data, whereas 
  \textbf{Non-blind} techniques require both the original cover media and data in order to perform data extraction. 
  \item \textbf{Fragility:} related to how the embedded data reacts to attacks and distortions applied to it. \textbf{Fragile} data are designed to show all attacks applied to it so that, when extracted, it is possible to verify which attacks have been applied to the media. This is useful when verifying the integrity of the media. \textbf{Semi-fragile} data are not robust against intentional distortions such as warping and noise filtering that attempt to degrade the embedded data, but are robust against content-preserving distortions such as compression and enhancement. Therefore, it can be used to trace any illegal distortions made to the media. 
  \item \textbf{Visibility:} whether the embedded data is visible to the human eye. 
  \item \textbf{Invertibility:} whether the embedded data can be removed from the cover media once embedded. Invertible data can be removed and non-invertible ones cannot. 
  \item \textbf{Robustness}: the ability of the embedded data to remain unchanged when attacks are applied to it. 

  \item \textbf{Security}: determines how difficult it is for an adversarial party to extract the data from the cover image. 
\end{itemize}

\section{Deep Learning-based Data Hiding Techniques}
\label{section:3}
Deep learning-based data hiding models utilise the encoder-decoder network structure to train models to imperceptibly and robustly hide information. They present an advantage over traditional data hiding algorithms because they can be retrained to become resistant to a range of attacks, and be applied to different end-use scenarios. Deep learning methods negate the need for expert knowledge when crafting data hiding algorithms, and improve security due to the black-box nature of deep learning models.  

\begin{figure}  \begin{center}
  \includegraphics[width=3.3in]{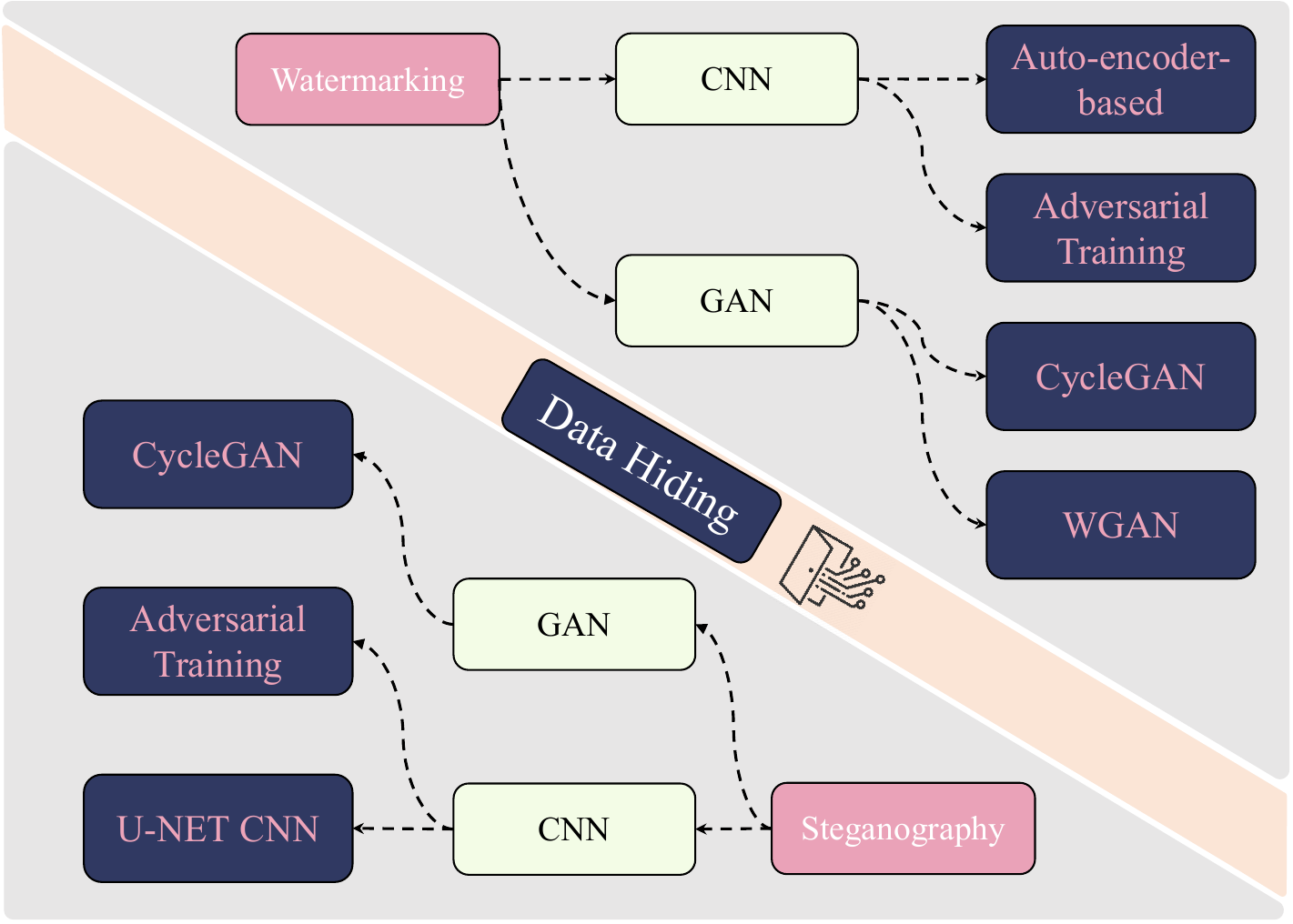}\\
  \caption{A hierarchical diagram showing the classification of deep learning-based data hiding models presented in this survey. `Adversarial training' refers to attack simulation during training, which includes noise-based attacks generated by a trained CNN.}\label{fig:diagram3}
  \end{center}
\end{figure}

The following section discusses deep learning-based data hiding techniques separated into techniques focused on watermarking and steganography. The detection and removal mechanisms are then discussed at the end of this section. 
The classification of deep learning-based data hiding techniques detailed in this section is outlined in Figure \ref{fig:diagram3}. It should be noted that CNNs incorporating adversarial training are different to GAN-based methods. Adversarial training in this instance refers to the use of trained CNNs for noise injection during the attack simulation stage, while GAN-based methods incorporate a discriminator to scrutinise encoded and cover images to improve embedding imperceptibility \cite{DA}. 

\begin{figure*}[t]
    \centering
    \includegraphics[width=0.94\textwidth]{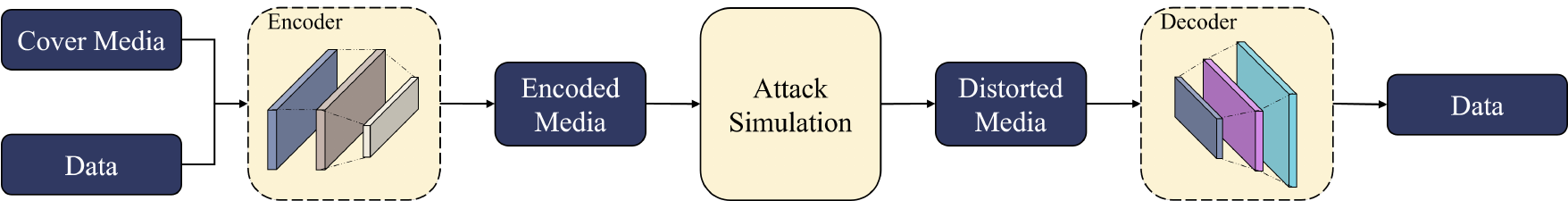}
    \caption{\centering{A diagram showing a general encoder-decoder architecture for digital watermarking.
    }}
    \label{fig:diagram4}
\end{figure*}

Currently, the majority of new watermarking models use an encoder-decoder architecture based on Convolutional Neural Networks (CNNs). A simple diagram showing the deep learning-based data hiding process can be found in Figure \ref{fig:diagram4}. 

In these models, the encoder embeds data in a piece of cover media; this encoded media is subjected to attack simulation, and then the data is extracted by the decoder network. Through the iterative learning process, the embedding strategy becomes more resistant to the attacks applied during simulation, and the extraction process improves the integrity of the extracted data. The advantage of this technique over previous traditional algorithms is that they require no expert knowledge to program, and can simply be retrained for different applications and attack types instead of needing to be designed from scratch. The system exists as a black box with high non-linearity where the intricacies of the embedding system are  unknown and impossible to ascertain. This makes deep learning-based methods highly secure, as well as adaptable to different end-use scenarios. Some variations of the simple CNN encoder-decoder approach shown above include convolutional auto-encoders, and CNNs with adversarial training components. The U-Net CNN architecture is common in steganography applications due to image segmentation abilities.

Many models adopt the Generative Adversarial Network (GAN) structure \cite{GAN}. The GAN framework consists of a generative model and a discriminative model. In deep data hiding, the discriminator network is given a mixture of encoded and unaltered images and must classify them as such. Throughout the learning process, the generative model improves in its data embedding capabilities, producing highly imperceptible examples, while the discriminative model improves at identifying encoded images. The end point of training is reached when the discriminator can only identify legitimately encoded images 50\% of the time -- it is making random guesses. The use of discriminative networks can greatly increase data imperceptibility, and is therefore useful for steganography as well as watermarking applications. 

There are also further variations of the GAN framework, including Wasserstein GANs (WGANs) and CycleGANs. 
The CycleGAN architecture is useful for image-to-image translation, and includes two generative and two discriminative models. The primary benefit of CycleGANs is that the model can be trained without paired examples. Instead, the first generator generates images from domain~A, and the second from domain B, where each generator takes an image from the other domain as input for the translation. Then, discriminator A takes as input both images from domain~A and output images from generator A, and determines whether they are real or fake (and vice versa for discriminator B). The resulting architecture is highly useful for translating between images.

\subsection {Deep Learning-based Watermarking Techniques}

In this section, current deep learning models for digital watermarking are categorised based on their network architecture design.

\subsubsection{Encoder-decoder Framework}
Due to the encoding and decoding tasks central to the data hiding process, the encoder-decoder deep learning framework is well suited for data hiding models. The encoder and decoder networks incorporate CNNs, which are used in a variety of applications such as detection, recognition, and classification due to their unique capabilities in representing data with limited numbers of parameters. The layers in the CNN learn non-linear, complex feature sets, representing the inputs and outputs to the network using weight-sharing mechanisms. The following deep watermarking models adopt the encoder-decoder framework without including a discriminator, which characterises GAN-based architectures \cite{firstCNN,DNN,wmnet,redmark,cnn15,DA}. A simple diagram of the encoder-decoder deep watermarking structure can be found in Figure \ref{fig:diagram4}. 

\noindent \textbf{Auto-encoder based Model. } 
The encoder-decoder architecture is a general framework and auto-encoder is a special case of the encoder-decoder structure. However, auto-encoder is usually used in unsupervised-learning scenarios by reconstructing the inputs. The potential of the CNN-based encoder-decoder frameworks for digital image watermarking was first explored in \cite{firstCNN}, which uses two traditional Convolutional Auto-Encoders for watermark embedding and extraction. Auto-encoder based CNNs were chosen based on their uses in feature extraction and denoising in visual tasks,  such as facial recognition and generation, and reconstructing handwritten digits. The intuition was that the auto-encoder CNN would be able to represent the watermark input in a form that was highly imperceptible when encoded within the cover image. This early CNN-based method applies two deep auto-encoders to rearrange the cover image pixels at the bit level to create a watermarked image. The technique was found to outperform the most robust traditional frequency domain methods in terms of both robustness and imperceptibility. Although promising for the future of deep watermarking, this technique is non-blind, and therefore not practically useful. 

Robust and blind digital watermarking results can be achieved using a relatively shallow network, as was shown by WMNet \cite{wmnet}. The watermarking process is separated into three stages: watermark embedding, attack simulation, where the CNN adaptively captures the robust features of various attacks, and updating, where the model’s weights are updated in order to minimise the loss function and thereby correctly extract the watermark message. Embedding is achieved by increasingly changing an image block to represent a watermark bit. The model is trained to extract watermark bits from the image blocks after attack simulations have been applied. 

The back-propagation embedding technique utilised in WMNet \cite{wmnet} used only a single detector network, which was found to cause performance degradation if the gradient computation in the backpropagation operation was affected by batch normalisation (BN). This deficiency was improved by adding an auto-encoder network as well as visual masking to allow flexible control of watermark visibility and robustness. The auto-encoder network was added to the encoder, and subsequently shorted time taken for both embedding and detection at the encoder because the feed-forward operation is generally much faster than back-propagation. These improvements were established in their follow-up work.

\noindent \textbf{Robustness Controls and Input Preprocessing.} 
Subsequent works after the aforementioned early techniques in \cite{firstCNN,wmnet} focused on generalising the watermarking process for multiple applications. Mechanisms such as robustness controls to influence the robustness/imperceptibility trade-off was introduced to gear models toward both watermarking and steganography applications, and mechanisms such as host and watermark adaptability were developed to pre-process inputs. 

A blind and robust watermarking technique was achieved using the CNN-based system~\cite{DNN}. The aim of this model is to generalise the watermarking process by training a deep neural network to learn the general rules of watermark embedding and extraction so that it can be used for a range of applications and combat unexpected distortions. The network structure is characterised by an invariance layer that functions to tolerate distortions not seen during network training. This layer uses a regularisation term to achieve sparse neuron activation, which enhances watermark robustness and computational efficiency. The layer also includes a redundancy parameter that can be adjusted to increase levels of redundancy in the resulting image, giving the model a higher tolerance of errors and increasing robustness. The primary aim of these features is to generalise watermarking rules without succumbing to overfitting. The model was compared with two auto-encoder CNN methods \cite{firstCNN,wmnet}, and was found to achieve greater robustness due to the new features it adopted. 

ReDMark \cite{redmark} uses two Full Convolutional Neural Networks (FCNs) for embedding and extraction along with a differentiable attack layer to simulate different distortions, creating an end-to-end training scheme. ReDMark is capable of learning many embedding patterns in different transform domains and can be trained for specific attacks, or against a range of attacks. The model also includes a diffusion mechanism  based on circular convolutional layers, allowing watermark data to be diffused across a wide area of an image rather than being confined to one image block. This improves robustness against heavy attacks, because if one image block is cropped out or corrupted, the watermark can still be recovered. The trade-off between robustness and imperceptibility can be controlled via a strength factor that can influence the pattern strength of the embedding network. 

The watermarking model developed by Lee \etal \cite{cnn15} uses a simple CNN for both embedding and extraction, without using any resolution-dependent layers. This allows for host image resolution adaptability – meaning that images of any resolution can be used as input to the system to be watermarked. There is an image pre-processing network that can adapt images of any resolution for the watermarking process. There is also watermark pre-processing, meaning the system can handle user-defined watermark data. This is achieved by using random binary data as the watermark that is updated at each iteration of training. The model also adopts a strength scaling factor, which allows for the controllability of the trade-off between robustness and imperceptibility. The method showed comparable, if not better, performance compared to ReDMark \cite{redmark}, and two generative adversarial-based models \cite{hidden,2stage}.  

\noindent \textbf{Adversarial Training.} A further improvement to the CNN-based encoder-decoder framework was made by adopting trained CNNs for attack simulation. While many other works use a fixed pool of attacks or a differentiable attack layer, using a trained CNN to generate attacks can greatly improve robustness, and introduces an adversarial component to model training. 
The focus of the Distortion Agnostic (DA) model \cite{DA} was to directly improve the HiDDeN model \cite{hidden}, primarily through adding robustness to the watermarking system in situations where the model is trained on a combination of distortions rather than one predetermined type. Instead of explicitly modelling different distortions during training from a fixed pool, the distortions are generated via adversarial training by a trained CNN. This technique was found to perform better in terms of robustness than HiDDeN \cite{hidden} when distortions not seen during training were applied to images. 
The DA framework also incorporates channel coding, a means of detecting and correcting errors during signal transmission, to add an additional layer of robustness to images by injecting extra redundancy. The watermark message is initially fed through the channel encoder to add redundancy before being input into the encoder model. Similarly, prior to extraction the redundant watermark message is input to a channel decoder to retrieve the final message. 

\subsubsection{Generative Adversarial Networks} 

The second primary approach for deep watermarking uses GANs, building upon the aforementioned techniques. Current models that adopt the GAN framework include \cite{hidden,romark,ABDH,IGA,2stage}. 
Many of the following models also use CNNs within their network architecture, but their use of the generative and discriminative components of the GAN framework set them apart from the aforementioned implementations. 

The first end-to-end trainable framework for data hiding was a model called HiDDeN \cite{hidden}, which uses an adversarial discriminator to improve performance. It was a highly influential paper that has informed the development of deep watermarking models since its release. The model consists of an encoder network, trained to embed an encoded bit string in a cover image whilst minimising perceptual perturbations, a decoder network, which receives the encoded image and attempts to extract the information, and an adversary network, which predicts whether or not an image has been encoded.

HiDDeN \cite{hidden} uses a novel embedding strategy based on adversarial examples. When neural networks classify image examples to a particular target class, invisible perturbations in the image can fool the network into misclassifying that example \cite{goodfellowadv}. These perturbations have been shown to remain preserved when exposed to a variety of image transformations \cite{kurakinadv}.  Adversarial examples are ordinarily a deficiency in neural networks since they reduce classification accuracy. However, since meaningful information can be extracted from imperceptible image perturbations, it was theorised that in a similar fashion meaningful information could be encoded in adversarial distortions and used as a watermark embedding strategy. This embedding technique, paired with the GAN framework, is able to achieve a higher payload capacity (measured in bits per pixel) than other common data hiding mechanisms such as Highly Undetectable Steganography (HUGO) \cite{hugo}, Wavelet Obtained Weights (WOW) \cite{wow}, and S-UNIWARD \cite{uniward}. One drawback of HiDDeN \cite{hidden} was that loss between encoded and decoded messages was minimised when trained only on a specific kind of attack compared to a combination of different attack types. This shows that the model is best when trained specifically to combat one type of attack, but not as effective when trained on a variety. 

This shortcoming was improved in the model ROMark \cite{romark}, which builds upon the framework from HiDDeN \cite{hidden} by using a min-max formulation for robust optimisation. This was done by addressing two main goals; first, to obtain the worst-case watermarked images with the largest decoding error, and second, to optimise the model’s parameters when dealing with the worst-case scenario so that decoding loss is minimised. The idea of this technique is to minimise decoding loss across a range of attacks, rather than training the model to resist specialised attacks, creating a more versatile and adaptable framework. Due to the optimisation for worst-case distortions in ROMark \cite{romark}, it performed better when trained on a combination of attacks, particularly on those that had not been seen during training. ROMark \cite{romark} was also more robust in some specialised attack categories, though HiDDeN \cite{hidden} had higher accuracy in these categories. 

Additional improvements were made to the framework from HiDDeN \cite{hidden} by Hamamoto et al. \cite{rotation}. This work uses a neural network for attack simulation rather than a single differentiable noise layer. There is a rotation layer followed by an additive noise layer, allowing the model to learn robustness against geometric rotation attacks. It also features a noise strength factor to control the robustness/imperceptibility trade-off. It was tested against HiDDeN \cite{hidden} and found to achieve greater image quality after watermark embedding, as well as greater robustness against JPEG compression. 
This model was the first to be trained to meet the Information Hiding Criteria (IHC) for robustness while simultaneously resisting geometric rotation attacks. It is suggested for future work to combine the architecture with a Scale Invariant Feature Transform (SIFT) detector \cite{sift}, which is able to detect image features that are robust for embedding to withstand geometric attacks in traditional watermarking algorithms.   

A novel embedding strategy using Inverse Gradient Attention (IGA) \cite{IGA} was adopted recently. Similar to Attention-based Data Hiding \cite{ABDH}, the focus is on identifying robust pixels for data hiding by using an attention mask. In the IGA method, the attention mask indicates the gradient values of the input cover image, which shows the robustness of each pixel for message reconstruction. It builds on the idea of adversarial examples first used for digital watermarking in HiDDeN \cite{hidden}, and uses the attention mechanism to locate the worst-case pixels for perturbation. By identifying robust pixel regions, this method further improves the payload capacity and robustness of watermarked images in tests against the following papers \cite{hidden, steganogan, DA}.   

A novel two-stage separable deep learning (TSDL) framework for watermarking is introduced by Liu \etal \cite{2stage}, which addresses the problems with one-stage end-to-end training (OET) such as slow convergence leading to image quality degradation, and having to simulate noise attacks using a differentiable layer. Instead of relying on differentiable approximations, the TDSL framework can use true non-differentiable noise attacks such as JPEG compression during training. This is because, in OET models, the encoder and decoder are trained using a differentiable noise layer, which means the noise must support backpropagation.  
The TDSL framework consists of noise-free end-to-end adversary training (FEAT), which is used to train an encoder that autonomously encodes watermarks with redundancy and without reference to any noise. Noise aware decoder only training (ADOT) is used to train the decoder so that it is robust and able to extract the watermark from any type of noise attack. The model was tested against HiDDeN \cite{hidden} and ReDMark \cite{redmark}, and found to achieve superior robustness for all but one attack category (crop). However, the two-stage training method is also robust against black-box noise attacks that are encapsulated in image processing software, which have not been tested in previous works. 

\noindent \textbf{Wasserstein GAN.}   
A popular variation of the traditional GAN technique is the Wasserstein GAN (WGAN) \cite{wgan}. This technique improves the model stability during training, as well as decreases the sensitivity of the training process to 
model architecture and hyperparameter configurations. The WGAN framework also provides a loss function that correlates with the quality of generated images. This is particularly useful for watermarking and steganography in the image domain, since image quality must be effectively optimised. Instead of the discriminator component of the network, WGANs include a critic. Rather than predicting the probability that a given image is real or fake, as is the discriminator's goal, the critic outputs a score denoting the `realness' of the input image. In a data hiding scenario, the encoder's aim is to maximise the score given by the critic for real instances – which corresponds to an encoded image. Current papers adopting the WGAN framework for their watermarking models include \cite{steganogan, spatialspread, double, texture}.   

Zhang \etal \cite{steganogan} introduced SteganoGAN. Three variants of encoder architecture are explored, each with different connectivity patterns. The basic variant applies two convolution blocks, where the encoded image is the output of the second block. Residual connections have been shown to improve model stability \cite{residual}, and in the residual variant, the cover image is added to the encoder outputs so that it learns to produce a residual image. The third variant uses a method inspired by DenseNet \cite{densenet}, in which there are additional feature connections between convolutional blocks that allow the feature maps generated by earlier blocks to be concatenated to those generated by subsequent blocks. 

Using adversarial training, this model achieves a relative payload of 4.4 BPP, 10 times higher than competing deep learning methods. Although both works \cite{hidden,steganogan} have mechanisms in place for handling arbitrarily-sized cover images as input, the higher payload capabilities of \cite{steganogan} means it can support a greater range of watermark data. The paper also proposes a new metric, Reed Solomon Bits Per Pixel (RS-BPP), to measure the payload capacity of deep learning-based data hiding techniques so that results can be compared with traditional data hiding methods. Although the primary focus for SteganoGAN is steganography, the high payload capacity and low detection rate of SteganoGAN produced images can also be applied to watermarking.    

Two further works by the same authors also use the WGAN framework. Plata \etal introduced a new embedding technique where the watermark is spread over the spatial domain of the image~\cite{spatialspread}. 
The watermark message is converted into a sequence of tuples, where the first element of each is converted to a binary representation. The spatial message is created by randomly assigning binary converted tuples to sections of the message, including redundant data for increased robustness. The paper also introduces a new technique for differentiable noise approximation of non-differentiable distortions which allows the simulation of subsampling attacks. The attack training pool is expanded from previous works to include subsampling and resizing attacks, and this wide range of attacks used during training increases general robustness. However, the spatial spread embedding technique reduces the embedding capacity, so is only useful for applications where capacity is not a priority. Furthermore, the training framework proposed requires half as much time as prior methods~\cite{hidden,redmark,DA}.  
The authors expand upon this work in a follow-up paper \cite{double}, which introduces the double discriminator-detector architecture. The discriminator is placed after the noise layer, and receives both noised cover images and noised encoded images, and thus the discriminator learns to distinguish watermarked and non-watermarked images with attacks already applied. In practical contexts, this is useful because it reduces the likelihood of false accusations being made of IP theft. If the image has already been attacked and must be proven to contain a watermark, this training technique is useful. Crucially, it does not degrade the overall robustness of the encoded images.

A technique for encoded image quality improvement is introduced by Wang \etal \cite{texture} based on texture analysis. The cover image texture features are analysed by a grey co-occurrence matrix which divides the image into complex and flat regions. The paper utilises the StegaStamp network~\cite{stegastamp} for embedding the watermark in the flat texture regions. This reduces the degree of image modification and improves the quality, and hence imperceptibility, of encoded images. The network in StegaStamp~\cite{stegastamp} produces higher quality images from low contrast examples;  therefore the contrast value is used to calculate the texture complexity of the image.

\begin{table*}
\centering
\caption{Summary table of deep learning-based digital watermarking methods.}
\label{tab_summary_watermarking}
\resizebox{\linewidth}{!}{
\begin{threeparttable}
\begin{tabular}{cccccccc} 
\toprule
\textbf{Arch} & \textbf{Model} & \textbf{Domain} & \textbf{Embedding Network} & \textbf{Extractor Network} & \textbf{RA} & \textbf{RC} & \textbf{Remarks}\\ 
\midrule
\multirow{2}{*}{AE CNN}
& \cite{firstCNN} & Frequency & AE CNN & AE CNN &  &  & First CNN for watermarking. Blind technique\\
& \cite{wmnet} & Spatial & AE \& RB & RB & & \checkmark & Autoencoder + visual mask\\
\midrule
\multirow{4}{*}{CNN}
& \cite{DA}& Spatial & Channel Coding \& CNN & Channel coding \& CNN & & & Channel Coding \& CNN for attacks\\ 
& \cite{redmark}& Frequency & CC layer \& DCT layer & DCT layer & \checkmark & \checkmark & Diffusion mechanism\\ 
& \cite{DNN} & Spatial & CNN & CNN & \checkmark & \checkmark & Invariance layer and image fusion\\
& \cite{cnn15} & Spatial & CNN \& AP & CNN & \checkmark & \checkmark & Robustness against geometric attacks\\
\midrule
\multirow{5}{*}{GAN}
& \cite{hidden} & Spatial & AL & GP \& CNN & & & Adversarial examples for embedding\\ 
& \cite{IGA}& Frequency & IG attention mask \& CNN & CNN & & & IG mask improves capacity and robustness\\ 
& \cite{romark} & Spatial & AL & FC layer & & \checkmark & Min-max formulation for robust optimisation\\
& \cite{2stage} & Spatial & AL & FC layer & \checkmark & \checkmark & Two stage training \\
& \cite{rotation} & Spatial & CNN & CNN + FC layers & & \checkmark & Robust against rotation and JPEG compression\\
\midrule
\multirow{4}{*}{WGAN} 
& \cite{steganogan} & Spatial & CB \& ASP & CNN & & & High payload capacity\\ 
& \cite{spatialspread} & Spatial & CNN & CNN + AP & & & Spatial spread embedding technique\\
& \cite{double} & Spatial & CNN & CNN + AP & & & Double discriminator-detector\\
& \cite{texture} & Spatial & CNN \& MP & CNN \& MP & & \checkmark & Uses texture analysis\\
\bottomrule
\end{tabular}
\begin{tablenotes}
\textbf{AE}: auto-encoder, \textbf{RB}: residual block, \textbf{AP}: average pooling, \textbf{ASP}: adaptive spatial pooling, \textbf{AL}: adversarial loss, \textbf{CB}: convolutional block, \textbf{FC}: fully connected, \textbf{IG}: inverse gradient, \textbf{GP}: global pooling, \textbf{MP}: max pooling, \textbf{CC}: circular convolutional, \textbf{RA}: resolution adaptability, and \textbf{RC}: robustness trade-off controls.
\end{tablenotes} 
\end{threeparttable}
}
\end{table*}

Attention-based Data Hiding (ADBH) \cite{ABDH} introduced an attention mechanism that helps the generative encoder  pinpoint areas on the cover image best suited for data embedding. The attention model is an input processing technique that allows the network to focus on specific aspects of a complex input one at a time. The attention model generates an attention mask, which represents the attention sensitivity of each pixel in the cover image. The value is regularised to a probability denoting whether altering the corresponding pixel will lead to obvious perceptual differences in the image that will be picked up by the discriminator. This improves the embedding process of the encoder network.

Rather than hiding a binary watermark message, this technique hides a secret image in the cover image. In the adopted CycleGAN framework, there is a target image generative model and a secret image generative model. The former generates watermarked images which are then fed to the cover image discriminative model, and builds on the PatchGAN  model to operate on one image patch at a time~\cite{patchgan}. Similarly, the secret image generative model serves as input for the secret image discriminative model. This architecture uses adversarial learning for both the embedding and extraction processes, ensuring that the distributions between both the cover and encoded images, and the encoded and extracted watermarks, are indistinguishable. The extractor network also operates on unmarked cover images, and inconsistent loss is used to make sure that, when this happens, there should be no correlation between the extracted data and the original watermark. 

The three central contributions of this paper -- the attention model, the use of cycle discriminative models, and the extra inconsistent loss objective -- were all tested in isolation and found to improve the performance of the model overall in terms of both image quality and robustness. 

\subsubsection{Discussion}
From the state-of-the-art deep learning methods discussed above, it seems that the GAN framework is the most promising in terms of robustness and secrecy optimisation due to the inclusion of adversarial loss in the objective calculations. To illustrate this, in HiDDeN \cite{hidden}, tests were conducted without including the adversary network and found to include perceptible alterations, whereas including the adversary greatly improved performance to produce an invisible watermarking technique. In tests comparing robustness, GAN-based models performed well, but were improved by techniques to target robust pixels for watermark embedding, such as the attention mechanisms used in Attention-based Data Hiding \cite{ABDH} and the IGA method \cite{IGA}. 

It is also important for models to be robust against a range of attacks. Papers such as those by Hamamoto \etal \cite{rotation} and Plata \etal \cite{spatialspread} incorporate geometric rotation techniques, and subsampling and resizing attacks respectively. Having a wider range of attack types during training increases general robustness. Additionally, using true non-differentiable noise, as shown in the two stage training technique developed by Liu \etal \cite{2stage}, provides better results for JPEG compression than models that use differentiable approximations. Using a trained CNN to generate attacks, Distortion Agnostic Watermarking \cite{DA} is also a promising approach for diversifying the attacks encountered during training. 

To produce an adaptable, generalised framework, it is important to include features such as host and watermark resolution adaptability, as well as robustness controls to influence the robustness/imperceptibility trade-off. Robustness controls make these models suitable for both watermarking and stenography, which relies more heavily on imperceptibility. The ideal digital watermarking model would include pre-processing networks for any watermark data or cover image to be input, as the framework was developed by Lee \etal \cite{cnn15}, while also taking advantage of an adversarial discriminator or critic. The techniques used in \cite{romark} to improve robustness by obtaining and optimising parameters for ROMark worst-case examples is also a promising technique for improving robustness.

If a future model could combine these features; attention mechanisms to improve embedding, robustness against geometric attacks, host and watermark resolution controls, robustness controls, and handling worst-case distortions, it could result in a highly robust and adaptable framework. However, the added overhead of all these added features could be an issue in practice.

Table~\ref{tab_summary_watermarking} shows a summary of the deep watermarking models reviewed in the above section. It shows whether the embedding strategy operates in the spatial or frequency domain, the nature of the watermark embedding and extracting networks, whether the technique supports host resolution adaptability (so that any cover image resolution can be used), and whether it includes controls for influencing the trade-off between imperceptibility and robustness. The remarks column describes any important information or novel contributions of the paper.

\subsection{Deep Learning-based Steganography Techniques}

This section classifies deep steganography techniques based on model architecture. Most techniques adopt the encoder-decoder structure shown in Figure \ref{fig:diagram4}, and are based on convolutional neural networks (CNNs). Many steganography implementations differ from watermarking implementations in their use of the U-Net structure for image segmentation \cite{unet}. Table \ref{tab_summary_steganography} shows a summary of deep learning-based steganography methods reviewed in this section.

\subsubsection{Encoder-decoder Framework}
CNN-based encoder-decoder structures have been adopted in the following papers \cite{hidingplain, stegnet, fusion, hidingdata, reversible, stegastamp, udh, longshortterm}. A simple diagram of the encoder-decoder deep watermarking structure can be found in Figure \ref{fig:diagram4}.

A paper from Google research \cite{hidingplain}, published in 2017, presented a deep steganography technique for hiding images inside other images. The structure contains three components in total. The first being the \textit{prep network}, which serves two purposes; to adjust the size of a smaller secret image to fit into the cover image, and to transform the colour-based pixels to recognisable features to be encoded into the cover image. The second layer of the encoder is called the \textit{hiding network}. As the name suggests, this layer creates the final stego-image. The final layer is the \textit{reveal network} which is used to decode the output from the second layer. The experiments in the work~\cite{hidingplain} were mainly conducted to show that it was possible to completely encode a large amount of data with limited visual disturbances in the cover media. However, such a technique lacked robustness, security, and was not of high quality. It was possible for attackers to recover both the cover and secret image with a trained network.   

Another image encoding technique called StegNet \cite{stegnet} was released in 2018. StegNet used structures from both auto-encoders and GANs to set up the encoding network. The cover image and the secret image were concatenated by a channel prior to the CNN encoding structure. Variance loss was included in loss calculations for the encoder and decoder. It was found that including the variance loss helped the neural network distribute the loss throughout the image rather than having concentrated areas of perceptual loss, improving the overall imperceptibility of embedding. The presented technique was highly robust against statistical analysis and when used against StegExpose \cite{stegexpose}, a commonly used steganalysis tool, it was also resistant against those attacks. Although robust, there were still some limitations to this method. Secret images and cover images must match in size and noise is still somewhat prominent in smoother regions.   

Comparing the method of Baluja et al. \cite{hidingplain} and StegNet \cite{stegnet}, there has been a vast improvement in image hiding. The inclusion of components adapted from GANs and auto-encoders in StegNet~\cite{stegnet} increased the robustness of the stego-images and was therefore more resistant to StegExpose~\cite{stegexpose}. Though StegNet \cite{stegnet} is quite robust, it is still lacking in some areas such as quality, image size restrictions and noise.   

A faster R-CNN (region-based CNN) method was introduced in \cite{fusion}. Firstly, the cover image is passed through a region proposal network, which makes the selection for feature extraction faster. Softmax loss is used to box these regions and then specific existing steganographic algorithms are selected and assigned to the boxed regions. Since \cite{fusion} uses a technique of selecting different steganography algorithms, using a combination of HUGO \cite{hugo}, S-UNIWARD \cite{uniward} and WOW~\cite{wow} algorithms, it is able to achieve highly imperceptible embedding. Being able to select effective areas on the cover image also allows Meng \etal \cite{fusion} to maintain a high level of robustness.   

The adaptation of the fusion technique \cite{fusion} allows for minimal distortion in the extracted stego-image. When looking at StegNet \cite{stegnet}, there is a concern for noise in smoother areas of the cover photo. Although \cite{stegnet} has been shown to be robust, it could be further improved in this area with the box selection the fusion method \cite{fusion} has. This does lead to a capacity dilemma if there was a method combining both StegNet \cite{stegnet} and the fusion method \cite{fusion}, since the latter would naturally be using less cover image area and therefore have a smaller capacity compared to StegNet \cite{stegnet}.    

A unique approach for steganography is shown by Sharma \etal in \cite{hidingdata}. In this paper, both the encoder and decoder consist of two layers. In the first layer, the \textit{prep layer}, smaller images are increased in size to correctly fit the cover image, there is a reconstruction of colour-based pixels to create more useful features, and the pixels are scrambled and then permutated. The second layer, the \textit{hiding layer}, produces the stego-image with the output of the first layer and the cover image inputted. The decoder consists of the \textit{reveal layer} and the \textit{decrypt layer}. The \textit{reveal layer} removes the cover image and the \textit{decrypt layer} decrypts the output of the \textit{reveal layer}. The advantage of this technique is that the first layer and its encryption method are similar to cryptography practices, allowing for a more secure embedded stego-image. Even when the cover media is known to the attacker, it is far more secure and difficult for the attacker to decode the secret image. This technique can also be applied to audio.    

A reversible image hiding method was introduced by Chang et al. \cite{longshortterm}. The structure relies on a concept called long short term memory. To encode, the cover image goes through a neural network in order to get a prediction, called the reference image. By subtracting the cover image from the reference image, the cover residuals (prediction errors) are calculated. Using histogram shifting (HS) on the cover residuals then produces stego residuals, along with an overflow map that is later used for the decoder. The stego-image is then created by adding the stego residual to the reference image. Where there is a pixel intensity flow, the overflow map is pre-calculated to flag these pixels. The decoder is essentially the reverse of the encoder, where the stego-image goes through a neural network to get a reference image, and the rest follows in reverse. This technique creates high quality, high capacity images that contain minimal noise. Its invertible feature to recover the cover image is unique to many other CNN based steganography.   

Tang et al. \cite{advemb} produced a steganography technique that uses adversarial embedding, called ADV-EMB. The network is able to hide a stego-message while being able to fool a CNN based steganalyser. The encoding network consists of a distortion minimalisation framework that adjusts and minimises the costs of the image according to features (gradients backpropagation) from the CNN steganalyser. The focus of ADV-EMB \cite{advemb} is to prevent steganalysers from being able to detect the stego-image. This shows in their results, with a high security rate and also increased imperceptibility. They are also able to train the system to counter unknown steganalysers by using a local well-performing CNN steganalyser as the target analyser, allowing for diverse applications. Although ADV-EMB \cite{advemb} is able to decrease the effectiveness of adversary-aware steganalysers, it has a weak pixel domain. However, an increase in payload capacity would increase the detection rate of steganalysers. 

\vspace{1mm}
\noindent \textbf{U-Net CNN.}
U-Net CNNs are used to facilitate more nuanced feature mapping through image segmentation. This is useful in image steganography applications because a cover image can be broken up into distinct segments based on a certain property (for example, \cite{reversible} uses a heatmap to score suitable embedding areas).


\begin{table*}
  \caption{Summary table of deep learning-based steganography methods.}
  \label{tab_summary_steganography}
    \begin{center} 
    \scalebox{0.78}{
    \begin{tabular}{ccccc} 
     \toprule
    \textbf{Arch} & \textbf{Model} & \textbf{Embedding Network} & \textbf{Extractor Network} & \textbf{Remarks} \\
    \midrule 
    \multirow{6}{*}{CNN}
    & \cite{hidingplain} & CNN & CNN & Embeds coloured image in coloured cover image \\
    & \cite{stegnet}  & CNN & CNN & Concatenating channel for embedding \\
    & \cite{fusion} & R-CNN \& CNN & N/A & Uses multiple techniques-based on selected regions \\
    & \cite{hidingdata}  & CNN & CNN & Embeds by scrambling and permutating pixels\\
    & \cite{longshortterm}  & LSTM w/ CNN \& HS & LSTM w/ Conv. layers \& HS & Uses HS for reversibility \\
    & \cite{advemb}  & CNN \& Adversarial Emb & N/A & Trains stego-images with steganalysers \\
    \midrule
    \multirow{1}{*}{CNN w/ Adv. Training}
    & \cite{securerobust}  & CNN with ReLU & CNN with ReLU & Several models for different priorities \\
    \midrule
    \multirow{2}{*}{U-Net CNN}
    & \cite{reversible}  & U-Net CNN w/ BN \& ReLU & CNN w/ BN and ReLU & Can withstand high and low frequencies \\
    & \cite{stegastamp}  & CNN U-Net & CNN & Embeds hyperlinks in images \\
    \midrule
    & \cite{udh}  & U-Net from Cycle-GAN & CNN & Improvement on \cite{hidden} for diverse use \\
    \multirow{3}{*}{GAN}
    & \cite{isgan}  & CNN & CNN & Hides data in Y channel of cover image \\
    & \cite{intermediatemedium}  & GAN \& CNN & N/A & Generates a textured cover image\\
    & \cite{coverless}  & CNN & CNN & Improvement on \cite{steganogan} in capacity and security \\
    \bottomrule
    \end{tabular}}
    \end{center}
\end{table*}

 A U-Net CNN technique for reversible steganography was developed by Ni et al. \cite{reversible}. This technique is capable of directly encoding the secret image into the cover image by concatenating the secret image into a six-channel tensor. Its decoder is formed from six convolution layers, each of which is followed by a batch normalisation and a ReLu activation layer. The embedding technique is not easily affected by excessively high or low frequency areas. It is also able to produce a high-quality image with a capacity that is in general better than other cover-selection and cover-synthesis based steganography techniques. Even with its high capacity capabilities, like other steganographic techniques that do not focus on robustness, too high of an embedding rate will increase the distortion rate more dramatically compared to robustness based steganography.

Universal Deep Hiding (UDH) is a model proposed for uses in digital watermarking, steganography, and light field messaging \cite{udh}. The encoder uses the simplified U-Net from Cycle-GAN \cite{cyclegan}, and a dense CNN for the decoder. The encoder hides the image in a cover-agnostic manner, meaning that it is not dependent on the cover image. UDH is an effective method due to the high-frequency discrepancy between the encoded image and the cover image. This discrepancy makes embedding robust in low-frequency cover images. UDH is also less sensitive to pixel intensity shifts on the cover image. The UDH method was compared with cover-dependent deep hiding (DDH). Since the encoding of the secret image was independent of the cover image, there was no method to adapt the encoding mechanism according to the cover image. The cover image may have some smoother areas which may not be ideal to embed data into, but with UDH there was no method of finding out this type of information. UDH is also unable to work well with severe uniform random noise. 

\noindent \textbf{CNN with Adversarial Training.}
The inclusion of adversarial attack networks during training can be used to help improve against steganalysis and promote robustness. Adversarial training helps the system distinguish small perturbations that an untrained steganography method may bypass. This is an important feature to have in steganography, since its main focus is to protect message security. 
Chen \etal developed a model that incorporates a trained CNN-based attack network that generates distortions \cite{securerobust}, similar to the technique used in the watermarking framework used by Xiyang \etal in Distortion Agnostic Watermarking \cite{DA}. The encoding structure is based off of a simple model that hides a secret grey-scale image into channel B (blue) of a coloured cover image, where both must have the same resolution. From this basic model, the paper was able to add two other enhanced models, a secure model and a secure robust model. The secure model inserts a steganalysis network into the basic model where its goal is to increase security against steganalysis. The secure and robust model uses the secure model and inserts an attack network to increase the robustness of the system. The separation of each model allows the framework by Chen \etal \cite{securerobust} to be used in several different scenarios, allowing the user to adjust to their needs. A downside is that users are unable to send RGB pictures and are limited to just grey-scale images. 

The secure model was shown to have the best invisibility against all models and was still the best when compared to the following methods \cite{hidingplain,endtoend}. The secure and robust model did not perform as well as the secure model but was still able to improve. The basic model and the secure model showed increased visual results in this comparison, with the secure model having the best visual results. The visual results of the secure and robust model were similar to ISGAN \cite{isgan}.   

\subsubsection{Generative Adversarial Networks based models}
Generative Adversarial Networks (GANs) are used extensively in deep steganography. Various new structures have allowed the simple GAN structure to be improved, increasing the effectiveness of steganography. It has brought up interesting techniques such as coverless steganography, and the ability to generate cover images. GAN-based architectures have been used in the following papers \cite{isgan, intermediatemedium, scyclegan, coverless, invertible}. 

ISGAN is a steganography technique that uses the GAN structure \cite{isgan}. Similarly to the secure technique developed by Chen \etal \cite{securerobust}, ISGAN is only able to send a grey-scale secret image. The encoder first converts the cover image into the YCrCb colour space where only the Y channel is used to hide the secret image since it holds luminance but no colour information. This colour space conversion does not affect backpropagation. The encoder also uses an inception module, which helps to fuse feature maps with different receptive field sizes. To aid the speed of training, the technique also adds a residual module and batch normalisation. A CNN is used to decode with batch normalisation added after every convolutional layer excluding the final layer. The results of this model showed that it was able to achieve a high level of robustness, with low detectability when scrutinised using steganalysis tools.   

In comparison to \cite{endtoend}, ISGAN residuals were less obvious, showing that the extracted secret image of ISGAN is much closer to the original than \cite{endtoend}. ISGAN is also able to achieve higher levels of invisibility since ISGAN uses a grey scale image. Thus, there is a trade-off between the complexity of the image (i.e. RGB values) and the imperceptibility of embedded information.   

Two separate designs are introduced by Li \etal in \cite{intermediatemedium}, which studies embedding data into texture images. The first model separates the texture image generation and secret image embedding processes. The texture image is generated using a deep convolutional generative neural network. The output of this is then used as the input for the concealing network for image hiding. The second model integrates the concealing network with the deep convolutional generative network. This second network should be able to generate the texture image while simultaneously embedding another image. The first model is easier to train and can be used in more diverse applications than the second model. Detection rates from steganalysis tools are almost 0 in both cases. These models provide high security, but with a few limitations. Firstly, the cover images generated are only textures and other subjects are not considered. Colour distortions also occur in the second model when the cover and secret images differ too much.   


Coverless steganography is possible because of the features of GAN. The general encoding idea of the proposed coverless method \cite{coverless} was that at first convolutional blocks were used to process the cover image to get a tensor, \textit{a}. The secret message was then concatenated to \textit{a} and processed through another convolutional block to get \textit{b} which was the same size as \textit{a}. The paper details two different models, one called the basic model and the dense model. The basic model uses the aforementioned encoding scheme, whereas the dense model includes a \textit{skip connection} to increase the embedding rate. The decoder for both models uses Reed Solomon algorithms on the tensor produced from the stego-image. The aim of this model was to improve the capacity and quality that other coverless steganography has not been able to achieve. Encoded image quality and payload capacity of the stego-images were improved when compared to \cite{steganogan}. The basic model introduced by Quin \etal~\cite{coverless} was able to perform significantly better in these aspects while the dense model was able only to match SteganoGAN \cite{steganogan}. The decoding network used in the coverless method \cite{coverless} was also more accurate than the one proposed for SteganoGAN~\cite{steganogan}. But there is a difference in the media encoded, where the coverless method \cite{coverless} encodes a string of binary messages while SteganoGAN \cite{steganogan} is able to encode an image. 

Many steganography techniques have not been invertible and leave the cover image distorted after removing the secret piece of media. With the method proposed by Chang \etal \cite{invertible}, the model was able to achieve invertible steganography. The method used is based on the Regular-Singular (RS) method. RS realises lossless data embedding through invertible noise hiding. There are three discriminate blocks used in this technique -- regular, singular and unusable -- in order to get the RS map. The adversarial learning component serves to capture the regularity of natural images. The model uses conditional GAN to synthesise, where the generator uses a U-Net structure and a Markovian discriminator is used. The use of GAN in conjunction with the RS method greatly improved upon the results of previous RS-based models.   

Currently, the paper \cite{reversible} uses a basic GAN structure and could be further improved or diversified with the adaption of other GAN structures. This could lead to further improvements in cover media recovery in terms of quality. Overall, the adversarial learning adopted in GAN-based models leads to greater robustness in steganography applications.

\noindent \textbf{CycleGAN.}
CycleGAN is a variation of the GAN architecture for image-to-image translation. The CycleGAN framework was adopted for steganography in S-CycleGAN \cite{scyclegan}. Within the model, there were three discriminators, two of which used the same function as the original CycleGAN framework. The third was an increased steganalysis module used to distinguish the stego-image from the generated images. The training cycle consists of three stages. First is the translation of the image from the X-domain to the style of the Y-domain. Second is the use of an LSB matching algorithm to embed the secret message into the output of the first stage. The third stage is where the stego-image is reconstructed to the input image of the first generator to the second generator. The full objective function included adversarial loss for all discriminators, as well as cycle consistency loss for the generative models. The advantage of S-CycleGAN is its ability to produce high quality images that are also robust against steganalysis. Overall S-CycleGAN \cite{scyclegan} showed results that were more resistant to detection, increasing the invisibility of embedded stego-images.   

When S-CycleGAN \cite{scyclegan} is compared against SGAN \cite{sgan}, the quality of the image is much higher, with 2.6x the Inception Score and 7x the Frechet Inception Distance of SGAN. In general, the results of S-CycleGAN \cite{scyclegan} showed that it was much more robust than SGAN, and the combined use of the original CycleGAN framework and traditional steganography algorithm S-UNIWARD \cite{uniward}.

\subsection{Data Hiding Detection and Removal Mechanisms}

Improved data hiding techniques have led to the development of more advanced strategies for detecting and removing hidden data. In the context of steganography, the field of steganalysis is devoted to identifying and extracting covert messages. Unlike steganography, watermarking is primarily intended for identifying the owner of the media rather than hiding secret messages. As a result, preventing the detection of data in watermarking is less critical, as long as the data cannot be modified or removed from the cover media. In watermarking, adversaries are more concerned with removing or degrading the watermark without altering the original cover media. It is important to understand the goals and methods of adversaries in data hiding depending on the application, since these scenarios should be mitigated by the data hiding strategy. 

\subsubsection{Steganlysis}
Steganalysis refers to the process of detecting covert steganographic messages from the perspective of an adversary. These techniques involve analyzing the media to identify any flaws in the embedding process and determine whether it has been encoded. There are two main categories of steganalysis techniques: signature steganalysis and statistical steganalysis, as noted in reference~\cite{s4}.

Signature steganalysis is comprised of two types called specific signature steganalysis and universal signature steganalysis. In specific signature steganalysis, the adversary is aware of the embedding method used, whereas universal signature steganalysis does not require this knowledge. Therefore, universal signature steganalysis can be used to detect several types of steganographic techniques \cite{s4,s16}.

The development of steganalysis, along with new steganography techniques, is crucial since it shows how robust new embedding techniques are to ever-improving detection technologies.

\subsubsection{Watermark Removal Strategies}

Watermark removal techniques can be separated into three categories \cite{security}. The first, blind watermark removal, is the technique that deep learning methods can defend against through varied attack simulation strategies. In blind removal techniques, the adversary has no knowledge of the watermarking process and attempts to degrade the watermark through attacks such as compression and geometric distortions. In key estimation attacks, the adversary has some knowledge of the watermarking scheme, and is then able to estimate the secret key used for embedding. Similarly, in tampering attacks the adversary has perfect knowledge of the watermarking scheme, resulting in a complete breakdown of the watermarking system where the secret key is explicitly obtained. In deep learning-based watermarking, the system is a black box even to its creators, hence the watermarking scheme cannot be uncovered by adversaries. Deep learning-based strategies only need to protect against the first type of attack, which is blind attacks. However, as watermark removal techniques improve through deep learning methods of their own, this is likely to change \cite{wmremoval}. 

\subsection{Results}

In this section, the robustness and imperceptibility of various deep data hiding models are presented and compared. Although there is no universally accepted standard dataset for testing deep data hiding models, COCO \cite{coco} is the most widely used dataset. Hence, the results achieved using this dataset were compared. Bit Error Rate (BER) and Peak Signal-to-Noise Ratio (PSNR) are the two commonly recorded metrics, although some papers only report one. It is important to note that the tests were conducted using different cover images and watermark dimensions, as indicated in the tables. Moreover, some papers employed a fixed pool of attacks during testing. Therefore, the results are not directly comparable but provide a general indication of the relative performance of the models.

Table \ref{tab:ber} compares deep data hiding models based on Bit Error Rate (BER), a measure of robustness. All models were tested using the COCO \cite{coco} dataset. The BER for all models is not directly comparable due to the different resolutions of cover images and the size of watermark payload. For example, a watermark embedded in a higher-resolution image can retain integrity without sacrificing encoded image quality, therefore improving robustness. Conversely, a smaller watermark payload can be embedded with less of a degradation in cover image quality, also improving robustness.  

\begin{table*}
  \caption{Comparing deep learning-based watermarking models based on their robustness, as measured by Bit Error Rate (BER). The `Results Origin' column indicates the source paper for each BER result.}
  \label{tab:ber}
    \begin{center}
    \scalebox{0.85}{
    \begin{tabular}{|c|c|c|c|c|c|} 
     \hline
    \textbf{Model} & \textbf{Results Origin} & \textbf{Architecture} & \textbf{Cover Image Dimensions} & \textbf{Watermark Bits} & \textbf{BER  (\%)} \\ 
     \hline
     HiDDeN \cite{hidden} & \cite{IGA} & GAN & 128 x 128 & 30 & 20.12\\
     \hline
     IGA \cite{IGA} & \cite{IGA} &  GAN & 128 x 128 & 30 & 17.88\\
    \hline
    DA \cite{DA} & \cite{IGA} & GAN &  128 x 128 & 30 & 20.48\\
    \hline
     ReDMark \cite{redmark} & \cite{redmark} & GAN & 128 x 128 & 30 & 18.18\\
     \hline
     Rotation \cite{rotation} & \cite{rotation} & GAN & 64 x 64 & 8 & 8.0\\
     \hline
     DNN \cite{DNN} & \cite{DNN} & CNN & 128 x 128 & 32 & 20.12\\
     \hline
     Double Detector-Discriminator \cite{double} & \cite{double} & GAN & 256 x 256 & 32 & 5.53\\
     \hline
     Spatial-Spread \cite{spatialspread} & \cite{double} & GAN & 256 x 256 & 32 & 8.38\\
    \hline
     Two-Stage \cite{2stage} & \cite{2stage} & GAN & 128 x 128 & 30 & 8.3\\
     \hline
    \end{tabular}}
    \end{center}
\end{table*}

Table \ref{tab:psnr} compares deep data hiding models based on Peak Signal-to-Noise Ratio (PSNR), a measure of encoded image quality. All models were tested using the COCO \cite{coco} dataset.

\begin{table*}
  \caption{Comparing deep data hiding models based on encoded image quality, measured using PSNR. The `Results Origin' column shows the paper which the PSNR result is taken from.}
  \label{tab:psnr}
    \begin{center}
    \scalebox{0.9}{
    \begin{tabular}{|c|c|c|c|c|c|} 
    \hline
    \textbf{Model} & \textbf{Results Origin} & \textbf{Architecture} & \textbf{Cover Image Dimensions} & \textbf{Watermark Bits} & \textbf{PSNR  (dB)} \\ 
     \hline
     Two-Stage \cite{2stage} & \cite{2stage} & GAN & 128 x 128 & 30 & 33.51\\
     \hline
     ABDH \cite{ABDH} & \cite{ABDH} &  CycleGAN & 512 x 512 & 256 & 31.79\\
    \hline
     ISGAN \cite{isgan} & \cite{ABDH} & GAN & 512 x 512 & 256 & 24.08\\
    \hline
     DA \cite{DA} & \cite{DA} & GAN & 128 x 128 & 30 & 33.7\\
     \hline
     HiDDeN \cite{hidden} & \cite{DA} & GAN & 128 x 128 & 30 & 32.3\\
     \hline
     DNN \cite{DNN} & \cite{DNN} & CNN & 128 x 128 & 32 & 39.93\\
     \hline
     WMNET \cite{cnn15} & \cite{DNN} & CNN & 128 x 128 & 32 & 38.01\\
     \hline
     ROMark \cite{romark} & \cite{romark} & GAN & 128 x 128 & 30 & 27.80\\
    \hline
     IGA \cite{IGA} & \cite{IGA} & GAN & 128 x 128 & 30 & 32.80\\
    \hline
     SteganoGAN \cite{steganogan} & \cite{IGA} & GAN & 128 x 128 & 30 & 30.10\\
     \hline
    \end{tabular}}
    \end{center}
\end{table*}

\section{Noise Injection Techniques}
\label{section:4}
Apart from model architecture, deep learning-based data hiding techniques can also be categorized based on the type of noise injection methods used. These methods are employed during training to simulate attacks and improve the robustness of the final model.

Several techniques have been utilized to enhance the robustness of deep learning-based data hiding models through attack simulation. The most frequently used approach involves employing a pre-defined set of attacks, which are simulated using a differentiable noise layer and subsequent geometric layers. Other methods include using a trained Convolutional Neural Network (CNN) to generate novel noise patterns, as discussed in references \cite{DA} and \cite{securerobust}, and adopting a two-stage training process to increase the complexity of the simulated attacks, as described in reference \cite{2stage}. The following section provides an overview of various noise injection techniques, along with their respective strengths and weaknesses.

\subsection{Identity}
Some earlier deep learning-based data hiding techniques \cite{firstCNN} did not involve attack simulation and were primarily intended as proof-of-concept demonstrations. Subsequent models generally include a control training scenario where no noise is added, such as the "Identity" layer in HiDDeN \cite{hidden}. It is unsurprising that techniques without attack simulation during training result in significantly lower levels of robustness. For instance, when subjected to JPEG and Crop distortions, the HiDDeN model trained exclusively with the Identity layer had only a 50\% success rate in successfully extracting watermarks and performed considerably worse when exposed to other types of attacks.

\subsection{Pre-defined Attack Pool} 
The most common technique for noise injection is to use a fixed pool of pre-defined attack types. Then, during training, the model can be exposed to one, multiple, or the entire range of attack types. Each attack includes an intensity factor that can be adjusted in order to vary the strength of the attack during training. In an ideal scenario, the model would be just as effective with all attack types when trained against the entire range as when trained with one specific attack. In most scenarios, the model uses a differentiable noise layer, meaning that all noise-based attacks must support back-propagation to cohesively work with the neural network \cite{hidden}. Since non-differentiable noise attacks such as JPEG compression are common in real attack scenarios, they are incorporated into training using differentiable approximations \cite{hidden}. The accuracy of these approximations is crucial to creating models that are robust against non-differentiable noise, therefore efforts have been made to revise and improve these approximations \cite{spatialspread}.  

Further improvements to the pre-defined attack pool include a greater range of attacks. For example, a rotation layer was added to the framework by Hamamoto et al. \cite{rotation}, and subsampling and resizing attacks were incorporated by Plata et al. \cite{spatialspread}. 
Various techniques have been introduced to improve the performance of models on a combined range of attacks, since early techniques using this method \cite{hidden} showed improved performance when trained with specific attacks rather than a range. For instance, ROMarK \cite{romark} adopts a technique wherein distortions that generate the largest decoding error are generated and used to optimise the model's parameters during training. 

\subsection{Adversarial Training} 
Another technique for noise injection does not use a fixed set of pre-defined attacks, but rather uses a separately trained CNN to generate novel noise-based distortions \cite{DA,securerobust}. The Distortion Agnostic model \cite{DA} generates distortions based on adversarial examples, using a CNN to generate a diverse set of image distortions. The Distortion Agnostic model was tested against HiDDeN and exposed to a number of distortions not seen during training, such as saturation, hue, and resizing attacks, and was found to be more effective. Although the Distortion Agnostic model could not surpass HiDDeN's effectiveness against a specific attack when trained only with that type, but this is not a practical training setup for an end use scenario. Using a trained CNN for attack generation can greatly improve robustness since distortions are generated in an adversarial training scenario. This means that distortions are generated explicitly to foil the decoder network throughout training. 

\subsection{Two-stage Training for Noise Injection}
A novel training method developed by Liu et al. \cite{2stage} improves noise injection techniques in two ways. Firstly, a training method dubbed Two-Stage Separable Deep Learning (TSDL) is used. This second stage of training is adopted for the decoder so that can work with any type of noise attacks, improving practicality. Because of this novel training scenario, this model can be trained using true non-differentiable distortion, rather than differentiable approximations. Aside from this distinction, a fixed pool of pre-defined attacks is still used during training. The use of true non-differentiable noise during training improved robustness against such attacks, including GIF and JPEG compression, but it is unclear from this work whether the increased overhead of the TDSL framework is a suitable trade-off for this improved performance, given the relative accuracy of differentiable approximations. 

\section{Objective Functions}
\label{section:5}

Objective functions are used when training machine learning models to optimise their performance. The function measures the difference between the actual and predicted data points. By optimising the loss function, the model iteratively learns to reduce prediction errors. The loss functions used for the discussed data hiding models fall into the regression category, which deals with predicting values over a continuous range, rather than a set number of categories. 

The aim of data hiding models is to optimise both robustness and imperceptibility, finding a balance between these two properties depending on the application's needs. The most common architecture for deep data hiding models is the encoder-decoder architecture where the model is partitioned into two separate networks for encoding and decoding. A common method for evaluating loss is to have two separate equations, one for the encoder and one for the decoder, and compute the total loss across the system as a weighted sum. Architectures that include an adversarial network also include adversarial loss in this sum. 
When optimising imperceptibility, the loss between the initial cover image and the final encoded image must be minimised. When optimising robustness, the loss between the initial secret message or watermark information and the final extracted message must be minimised. This is the bases for the encoder-decoder loss method.  
This section will go through some common objective functions utilised in deep learning models for data hiding, as well as novel strategies that do not follow the encoder-decoder loss framework.

\subsection{Mean Squared Error (MSE)} 
A function used to compute the difference between two sets of data. It squares the difference between data points in order to highlight errors further away from the regression line. In image-based data hiding, it can be used to compare the cover image to the encoded image by taking the square of the difference between all pixels in each image and dividing this by the number of pixels. The equation's sensitivity to outliers makes it useful when pinpointing obvious perceptual differences between two images. For example, a cluster of pixels that are obviously different from the original is perceptually obvious, but a change distributed over all image pixels is harder to spot. The equation is also used to compute the difference between the original and encoded watermark in some applications. 
For two images $X$ and $Y$ of dimensions W $\times$ H, the mean square error is given by the following equation:
\begin{equation} \label{eq: 1}
MSE(X,Y) = \frac{1}{WH}\sum_{i=1}^{W}\sum_{j=1}^{H}(X_{i,j} - Y_{i,j})^2.
\end{equation}

MSE is used in these works~\cite{cnn15,steganogan,rotation,spatialspread,double,coverless,isgan,intermediatemedium} to calculate loss at the encoder, while many works~\cite{2stage,redmark,hidingdata,stegnet,reversible,securerobust,hidden,DA,texture,stegastamp} use it to compute both encoder and decoder loss. 
MSE is used in conjunction with SSIM \ref{eq: 17} to calculate perceptual difference at the encoder~\cite{isgan,intermediatemedium}.
The papers \cite{texture,stegastamp} also use LPIPS perceptual loss \cite{lpips} to evaluate loss at the encoder, which is discussed in Section \ref{section:5.2}.
The Distortion Agnostic Model \cite{DA} incorporates additional adversarial loss into its Euclidean Distance functions. Loss at the encoder incorporates GAN loss with spectral normalisation to further control the quality of the encoded image. It is similar to the adversarial loss function used in \cite{hidden} in Equation~\eqref{eq: 6}, but is incorporated into the encoder loss function rather than being weighted separately in the overall weighted sum.   

\subsection{Mean Absolute Error (MAE)} Similar to MSE, MAE is used to compute the difference between two sets of data, but instead takes the absolute value of the difference between data points. Therefore, it does not highlight errors further from the regression line or give any special significance to outliers. It is used to compute loss at the decoder in \cite{cnn15}. Since the secret message or watermark consists of binary values, it is more suitable to use MAE, which is suited to discrete values, to compute the difference to ensure balanced training of both the encoder and decoder networks. Additionally, it is used to calculate encoder and decoder loss in \cite{invertible,longshortterm,stegnet,udh}.
For two binary watermarks $M$ and $M'$ of dimensions $X$ x $Y$, the mean square error is given by the following equation:
\begin{equation} \label{eq: 2}
MAE(M,M') = \frac{1}{XY}\sum_{i=1}^{X}\sum_{j=1}^{Y}|M{(i,j)} - M'{(i,j)}|
\end{equation}

\subsection{Cross Entropy Loss} This function is used to measure the difference between two probability distributions for a given variable, in this case the initial and extracted binary watermark message. The equation takes in $p(x)$, the true distribution, and $q(x)$, the estimated distribution. In digital watermarking this corresponds to the original watermark and the extracted watermark respectively, where $q(x)$ is the output of the decoder network representing the probability of watermark bits. Cross entropy loss takes the log of the predicted probability, therefore, the model will be `punished' for a making a high probability prediction further from the true distribution. Cross entropy loss is given by the following equation:
\begin{equation} \label{eq: 5}
 H(p,q) = - \sum_{x}p(x)log(q(x)).  
\end{equation}
This technique is used for measuring loss at the decoder in \cite{steganogan,redmark,texture,rotation,stegastamp}, and used to compute adversarial loss in \cite{advemb,invertible}  

\subsection{Mean-Variance Loss}
Variance loss is used to calculate how loss varies across an entire distribution. Using variance loss in conjunction with mean calculations leads to loss being distributed throughout an image rather than concentrated in certain areas. When calculating loss at the encoder, it is important that there are no concentrated areas of difference, since this will be easily perceptible. The paper \cite{stegnet} adopts variance loss and MAE loss for both encoder and decoder loss calculations.

A similar technique is applied by \cite{spatialspread,double}, but for a different reason. Variance loss is used to calculate loss at the decoder rather than MSE since the redundant data used to create the watermark message means the original watermark and extracted watermark need not be identical. Rather than extract this redundant data, as was done in \cite{cnn15}, the loss function was modified to a joint mean and variance function. 
For original message $M$ and extracted message $M'$, the mean and variance are given by:

\begin{equation} 
L_m = (M, M') = \frac{b^2}{HW(n+k)}||M-M'||_1,
\end{equation}

\begin{equation} 
L_v = (M, M') = \frac{b^2}{HW}\sum^{H_b}_{h=0}\sum^{W_b}_{w=0}Var(|M-M'|),
\end{equation}
where $b$ is the number of times every `slice' of $M$ is replicated across the image in the spatial spread embedding technique \cite{spatialspread}, and $k$ relates to the number of tuples in the sequence created in this technique.

\subsection{Adversarial Loss}

For deep learning models that use a discriminator network for adversarial learning, an additional loss function is added to the overall optimisation that accounts for the discriminator's ability to detect an encoded image. 
Adversarial loss can be combined with loss at the encoder, since the encoder is the generative network, to improve the imperceptibility of the encoded watermark. The discriminator takes in an image $X$ that is either encoded or unaltered, and predicts $A(I)\epsilon[0,1]$, denoting the probability that $X$ has been encoded with a watermark. The discriminator optimises the function for adversarial loss from its predictions given by the following equation:
\begin{equation} \label{eq: 6}
l_a(X, Y) = log(1-A(I_c)+log(A(Y)),  
\end{equation}
where $X$ is the cover image, $Y$ is the encoded image. This equation is combined with the loss function for encoder loss in \cite{hidden,romark,rotation}.   

\subsection{KL Divergence}
KL divergence is used to quantify the difference between probability distributions for a given random variable, for instance, the difference between an actual and observed probability distribution. It is often used in Generative Adversarial Networks to approximate the target probability distribution. This equation was used in \cite{coverless} to calculate adversarial loss, where it is used in conjunction with JS divergence. This method for calculating adversarial loss can be represented in the following way: 

\begin{equation} \label{eq: 7}
    KL(P||Q) = \sum_{c\epsilon{C}}P(c)log\frac{P(c)}{Q(c)},
\end{equation}
\begin{equation}
    JS(P||Q) = \frac{1}{2}KL(P||\frac{P+Q}{2})+\frac{1}{2}KL(Q||\frac{P+Q}{2}),
\end{equation} \label{eq: 8}
where $c$ is the cover image, $P$ is the probability distribution function for all cover images, and $Q$ is the probability distribution function for the generated steganographic images. 

\subsection{Cycle Consistency Loss} This function is often used for CycleGAN implementations. The function performs unpaired image-to-image translation and is useful for problems where forward and backwards consistency between mapping functions is important.
In both \cite{ABDH,scyclegan}, there are two generative and two discriminative models used, therefore the framework needs to learn the bijective mapping relationship between two image collections, where the first contains the original cover images and the second contains the secret images that serve as watermarks. The generative model needs to learn to transform from one source image domain to another, and this mapping relationship cannot ensure that the extracted watermark is the same as the original. 
Cycle consistency loss is detailed in \cite{scyclegan} as an equation for the transformation of an X-domain style image to a Y-domain style image, where the generator models for both domains satisfy backward cycle consistency. Cycle adversarial loss is abstracted as the following:
\begin{equation} \label{eq: 9}
    D_{M}(M,M')=G_S(G_e(X,M,A)),
\end{equation}
where $D_S$ represents the cycle discriminative network judging the extracted watermarks, $M$ is the original secret watermark, $M'$ is the extracted watermark, $G_S$ is the generative model that generates secret images to use as watermarks, $X$ is the cover image, and $A$ is the attention mask.   

\subsection{Wasserstein Loss}

The following models \cite{steganogan,spatialspread,double,texture,stegastamp} use a WGAN framework, therefore using a critic network rather than a discriminator to train the encoder. Rather than classifying an example as ‘real’ or ‘fake’, the critic outputs a number. The aim is to maximise this number for ‘real’ instances, which in this case are images that have been encoded with a message. Wasserstein loss is given by the following equation:
\begin{equation} \label{eq: 10}
l_w = C(X) - C(E(X, M)),
\end{equation}
where $C(X)$ is the critic output for the initial cover image, and $C(E(X,M)) $ represents the critic output for the encoded image $E$ consisting of the cover image and watermark message $M$. 

\subsection{Novel Approaches}

Certain approaches to performance optimisation do not follow the above framework for optimising at the encoder and decoder, along with a third possible adversarial component. 

\vspace{1mm}
\noindent \textbf{Image Fusion.} 
The technique developed in \cite{DNN} differs from many other deep learning techniques for digital watermarking by approaching the process as an \textit{image fusion} task, aiming to maximise the correlation between the feature spaces of images. 

The previous techniques focus on preserving certain parts of the cover image in the resulting encoded image. Different optimisations are applied to control embedding and extraction, and weights are used to control watermarking strength. Conversely, \cite{DNN} takes 2 input spaces, $C$ denoting the cover image, and $W$ denoting the watermark. $W$ is mapped onto one of its latent feature spaces $W_f$. Watermark embedding is performed by the function $\{W_f,C\} \to M$, where $M$ denotes the fusion feature space of the watermark and cover image, creating an intermediate latent space. 
In this system, the loss is computed using the \textit{correlation} function given by:

\begin{equation} 
\begin{aligned}
\psi(m_i,w_f^i)=\frac{1}{2}(||g(f_1(w_f^i)),g(f_1(m_i))||_1+\\
||g(f_2(w_f^i)), g(f_2(m_i))||_1), 
\end{aligned}
\label{eq: 11}
\end{equation}

where $m_i$ and $w_f^i$ are examples of the spaces $M$ and $W$ respectively. $g$ denotes the Gram matrix of all possible inner products and $f$ represents the convolutional blocks in the encoder network. A similar image fusion approach developed independently is utilised in \cite{fusion}  

\vspace{1mm}
\noindent \textbf{Distortion Minimisation Framework.} The steganography model is detailed in \cite{advemb}, steganography is formulated as an optimisation problem with a payload constraint:

\begin{equation}
    min_SD(X,Y), \psi(Y)=k, 
\end{equation}
where $X$ is the cover image, $Y$ is the encoded stego-image, and $D(X,Y)$ is a function measuring the distortion caused by modifying $X$ to create $Y$. $\psi(S)$ represents the data extracted from $Y$ and $k$ is the number of bits that make up the data. Different additive distortion functions may be used for $D$ that measure the cost of increasing each pixel value in the cover image by 1. Large cost values are assigned to pixels more likely to cause perpetual differences in the final stego-image, and will therefore have a low probability of being modified during embedding. 

\vspace{1mm}
\noindent \textbf{Inconsistent Loss.} The CycleGAN model is introduced in \cite{ABDH}, there is an additional loss function called inconsistent loss that ensures that a secret watermark image can only be extracted from an encoded image. Since the second generative model $G_S$ that extracts the watermarks also receives unencoded cover images as input, any data it extracts should be completely different from the real secret watermark image. This is given by the following equation:
\begin{equation}
    max_{G_S}(G_S,M') = max_{G_S}|G_S(X) - G_S(Y)|,
\end{equation} \label{eq: 12}
where $M'$ is the extracted watermark image, $X$ is the original cover image, and $Y$ is the encoded `target' image. \cite{ABDH} combines this with adversarial loss and cycle adversarial loss to evaluate the model's overall performance.   

\section{Evaluation Metrics}
\label{section:6}

Evaluation metrics for data hiding techniques can be divided into two primary categories; those evaluating robustness and those evaluating encoded image quality. Additionally, there are metrics for measuring information capacity. 

\subsection{Robustness}

The following metrics are used to evaluate the robustness of the embedded data. After attacks are applied to the encoded image and the data is extracted, these metrics are applied to compute the difference between the original message data and the data extracted from the image to judge how well that data survived distortions applied to it. 

\subsubsection{Bit Error Rate (BER)} is the number of bit errors divided by the total number of bits transferred over a certain time frame. It therefore gives the percentage of erroneous bits during the transmission. It is used to compare the extracted message and the initial message, each converted into binary. 
The BER between the original message $X$ and extracted message $Y$ is given by the following equation:

\begin{equation} \label{eq: 13}
BER(X,Y) = 100\,\frac{\text{{\small\#}\;Bits in error}}{\,\text{{\small\#}}_{Total}\; \text{Bits transmitted}}
\end{equation}

\subsubsection{Normalised Correlation (NC)} is a measure of the similarity between two signals as a relative displacement function.  
Normalised correlation gives values in the range [0,1] with a higher value denoting a higher similarity between images. The normalised correlation between the original message $X$ and extracted message $Y$ is given by the following equation: 
\begin{equation} \label{eq: 14}
NC(X,Y) = \frac{1}{WH}\sum_{i=0}^{W-1}\sum_{j=0}^{H-1}\delta(X_{i,j},Y_{i,j}), 
\end{equation}
where \[ \delta({X,Y})= 
    \begin{cases}
      1, & \text{if}\ X=Y \\
      0, & \text{otherwise}
    \end{cases}
    \]

\subsection{Quality}
\label{section:5.2}

The following metrics are used to evaluate the quality of the encoded image. It is linked to the property of imperceptibility since embedded data should not produce any detectable perturbations in the image. 

\subsubsection{Peak Signal to Noise Ratio (PSNR)} computes the difference between two images by taking the ratio between the maximum power of the signal and the power of corrupting noise that affects the signal. In data hiding, it is used to evaluate the difference between the cover image and the encoded image, showing the effectiveness of the embedding technique. PSNR is expressed in decibels (dB) on a logarithmic scale, with a value of above 30dB generally indicating that the image difference is not visible to the human eye. Therefore, PSNR can be used to quantify data imperceptibility. 
PSNR is defined by taking the Mean Square Error (MSE) of two images (cover image $I_c$ and watermarked image $I_w$) using the following equation. 
\begin{equation} \label{eq: 16}
PSNR(I_c, I_w) = 10\, log_{10}\left(\frac{255^2}{MSE(I_c, I_w)} \right) 
\end{equation}

\subsubsection{Structural Similarity Index (SSIM)} a perception-based model that measures image degradation as perceived changes in structural information. It considers variance and covariance, which measure the dynamic range of pixels in an image. Unlike MSE and PSNR, which measure absolute error, SSIM considers the dependencies between spatially close pixels, incorporating luminance and contrast masking. SSIM is given in the range [-1.0,1.0], with 1 indicating two identical images. 
The SSIM of two images is denoted as follows, where the mean of the two images is given by $\mu_x$ and $\mu_y$, and their variances are given by $\sigma^2_x$ and $\sigma^2_y$.
\begin{equation} \label{eq: 17}
 SSIM(x, y) = \frac{(2\mu_x\mu_y + k_1)(2\sigma_x\sigma_y+k_2)}{(\mu^2_x+\mu^2_y+k_1)(\sigma^2_x+\sigma^2_y+k_2)} 
\end{equation}
The constants $k_1$ and $k_2$ exist to stop a 0/0 calculation. In the default configuration, $k_1 = 0.01$ and $k_2 = 0.03$.

\subsubsection{Learned Perceptual Image Patch Similarity (LPIPS)} is another type of perceptual loss. The aim of LPIPS \cite{lpips} is to measure human perceptual capabilities by going beyond the simpler SSIM \ref{eq: 17} and PSNR \ref{eq: 16} metrics, studying the deeper nuances of human perception. It is a model that evaluates the distance between image patches. The model uses a large set of distortions and real algorithm outputs. There is consideration of traditional distortions, noise patterns, filtering, spatial warping operations and CNN-based algorithm outputs. LPIPS is able to perform better due to a larger dataset used compared to other datasets similar to this kind, as well as being able to use the outputs of real algorithms. LPIPS has a scoring system where the lower the score, the more similar to the compared image while a higher score indicates a larger difference between the images. 

\subsubsection{Frechet Inception Distance (FID)} is a model that measures the quality of a synthetic image. It was mainly developed to check the quality and performance of images generated by GAN structures. FID uses the inception v3 model \cite{inception} where in the last layer of FID, it is used to get the computer-vision-specific features of an input image. They are calculated for the activations for all images, real and generated. These two distributions are then calculated using FID.
The score compares the quality of synthetic images based on how well inception v3 classifies them as one of the 1,000 known objects. Essentially the generated images are not compared to real images, instead, they are compared to other synthetic images that have already been compared to real images and scoring their similarity.
A low FID score indicates a higher quality image while a high FID score indicates a lower quality image. 

\subsection{Capacity}

Capacity is a measure of the amount of information that can be included in the data payload and subsequently embedded into the cover media. 

\subsubsection{Bits Per Pixel (BPP)} The number of bits is used to define the colour value of a pixel. A higher BPP value denotes a higher number of possible colour values for a pixel. Therefore, a higher BPP value for payload capacity shows that more identifying data can be embedded in the pixels of an image. BPP is used to calculate data payload capacity in \cite{steganogan}. 

\subsubsection{Reed Solomon Bits Per Pixel (RS-BPP)} The paper~\cite{steganogan} introduces a new metric RS-BPP based on Reed-Solomon codes. In deep data hiding models, measuring the number of bits that can be embedded per pixel depends on the model, the cover image, and the model itself, and is therefore non-trivial to measure just in BPP. 
Given an algorithm that returns an erroneous bit with probability $p$, the aim is to have a number of incorrect bits that is less than or equal to the number of bits that can be corrected. 
This can be represented by the equation:
\begin{equation} \label{eq: 15}
    pn \leq \frac{n-k}{2}, 
\end{equation}
where the $\frac{k}{n}$ ratio is the number of data bits that can be transmitted for each bit of the message. RS-BPP allows deep data hiding techniques such as \cite{steganogan} to be directly compared to traditional algorithms in terms of capacity. 

\section{Datasets} 
\label{section:7}

This section presents a table that displays the image databases utilized to train the deep data hiding models discussed in this survey.

\vspace{1mm}
\noindent \textbf{Describable Textures Dataset (DTD):} The DTD dataset includes multiple  kinds of labelled texture images, totalling 5640 images, and was used to train~\cite{intermediatemedium}.

\begin{table*}
  \caption{Summary table of datasets used to train deep learning-based data hiding models.}
  \label{tab:dataset}
    \begin{center} 
    \scalebox{0.73}{
    \begin{tabular}{|c|c|c|} 
     \hline
    \textbf{Name} & \textbf{Trained} & \textbf{Description} \\
    \hline 
    COCO \cite{coco} & \cite{hidden,steganogan,IGA, ABDH, DA,rotation,spatialspread,double,intermediatemedium,fusion} & 330 images of everyday scenes. Cluttered images useful for DH \\
    \hline
    DIV2K \cite{div2k} & \cite{steganogan,IGA,coverless} & 1k low resolution images. Includes open scenery difficult for DH \\
    \hline
    CIFAR-10 \cite{cifar} & \cite{redmark, DNN} & 60k 32x32 images of singular objects \\
    \hline
    Pascal VOC \cite{pascal} & \cite{redmark,isgan} & 15k images of singular objects \\
    \hline
    BOSSBase \cite{boss} & \cite{wmnet, cnn15, invertible, advemb,longshortterm} & 9k 512x512 greyscale images in PGM format \\
    \hline
    ImageNet \cite{imagenet} & \cite{DNN,isgan,scyclegan,advemb,securerobust,hidingplain,stegnet,reversible,udh, stegastamp} & 14mil images organised based on WordNet hierarchy \\
    \hline
    MIRFLICKR \cite{mirflickr} & \cite{texture,stegastamp} & 1mil Flickr images under Creative Commons license\\
    \hline
    Flickr30k \cite{flickr30} & \cite{hidingdata} & 31k images from Flickr under Creative Commons license\\
    \hline
    Labeled Faces in the Wild (LFW) \cite{LFW} & \cite{isgan} & 13k images of faces from the web\\
    \hline
    Describable Textures Dataset (DTD) \cite{textureds} & \cite{intermediatemedium} & 5k labelled texture images\\
    \hline
    \end{tabular}}
    \end{center}
\end{table*}

\vspace{1mm}
\noindent \textbf{COCO:} COCO is a large-scale object detection, segmentation, and captioning dataset \cite{coco}. The dataset contains 330 thousand images of complex, everyday scenes including common objects. The goal of the dataset was to advance AI scene understanding by combining the more narrow goals of image classification, object localisation, and semantic segmentation. Since the dataset consists of relatively cluttered images picturing multiple objects, there are more surfaces and variations in textures in which to embed watermarking data, making it useful and popular for data hiding applications. COCO was used to train \cite{hidden,steganogan,IGA, ABDH, DA,rotation,spatialspread,double,intermediatemedium,fusion} and to test \cite{redmark,DNN,2stage,romark}.

\vspace{1mm}
\noindent \textbf{DIV2K:} Div2K is a single-image super-resolution dataset of 1000 images of different scenes. The dataset consists of low resolution images with different types of degradations applied. Compared to COCO, it contains more images of open scenery, which can present a challenge for data embedding. Therefore, it is used to train \cite{steganogan,IGA} in conjunction with \cite{coco} to ensure the model is trained on a variety of different types of images. On its own, it is used to train \cite{coverless}.

\vspace{1mm}
\noindent \textbf{CIFAR-10:} CIFAR-10 is a labelled subset of the 80-million Tiny Images Dataset \cite{cifar}, and contains 60 thousand small 32x32 images. They mostly picture single objects such as animals or vehicles, and are separated into 10 classes depending on the subject, though these are ignored for data hiding applications. CIFAR-10 was used to train \cite{redmark, DNN}.

\vspace{1mm}
\noindent \textbf{Pascal VOC:} Similar to \cite{cifar}, the Pascal VOC dataset \cite{pascal} includes pictures of objects such as vehicles and animals rather than full scenes with multiple objects. This dataset has been widely used as a benchmark for object detection, semantic segmentation, and classification tasks. It was used to train \cite{redmark,isgan}.

\vspace{1mm}
\noindent \textbf{BOSSBase:} This dataset contains 9074 512x512 greyscale images in PGM format \cite{boss}. It is widely used for training steganography algorithms. The dataset was used to train \cite{wmnet, cnn15, invertible, advemb,longshortterm}. 

\vspace{1mm}
\noindent \textbf{ImageNet:} 
ImageNet contains 14 million images organized according to the WordNet hierarchy~\cite{imagenet}. Images are classified into `synsets' based on their subjects, all sorted and labelled accordingly. It was used to train \cite{DNN,isgan,scyclegan,advemb,securerobust,hidingplain,stegnet,reversible,udh, stegastamp}.

\vspace{1mm}
\noindent \textbf{MIRFLICKR:}
The MIRFLICKR dataset \cite{mirflickr} contains 1 million Flickr images under the Creative Commons license. It is used to train \cite{texture,stegastamp}. 

\vspace{1mm}
\noindent \textbf{Flickr30k:} The Flickr30k dataset \cite{flickr30} contains 31,000 images collected from Flickr, together with 5 reference sentences provided by human annotators. It was used to train \cite{hidingdata}. 

\vspace{1mm}
\noindent \textbf{Labeled Faces in the Wild (LFW):} The LFW dataset \cite{LFW} public benchmark for face verification, contains more than 13,000 images of faces collected from the web. The dataset was used to train~\cite{isgan}.

\section{Open Questions and Future Work}
\label{section:8}

Deep learning for data hiding is a new and evolving research field, and there remain many different avenues to consider moving forward. This section will discuss some open questions that we believe warrant further research and consideration including expanding the applications of digital watermarking to other media domains, pursuing deep learning-based language watermarking, improving robustness against deep learning-based watermark removal attacks, watermarking machine learning models, combating the use of watermarking to launch backdoor attacks on machine learning models, and exploring the applications of watermarking for detecting and identifying synthetic media. Finally, future directions for steganography are discussed, including its potential use for spreading malware.

\subsection{Expanding Applications for Deep Learning Digital Watermarking Models}

While the deep watermarking models discussed in this survey were primarily focused on image watermarking, there is significant potential for applying watermarking to other types of media. Although there are many traditional algorithms focused on watermarking video, audio, 3D models, and electronics, there is yet to be any deep learning models focusing on these areas. For instance, various techniques have been proposed for watermarking 3D models \cite{3dsurvey2}, including new promising methods for watermarking vertices data \cite{3dtrad}. There are also existing works concerning GANs generating 3D models \cite{gan3dgen,3dsurvey}. Furthermore, a recent paper from Google \cite{mesh} details a promising deep learning technique of embedding watermark messages into simple 3D meshes which can then be decoded from 2D rendered images of the model from multiple angles and lighting configurations. It is noted in this work that robustness and capacity will need to be improved before practical application, particularly robustness to non-differentiable 3D attacks. More complex models and lighting arrangements could be explored with a better-quality renderer. A paper from Google research by Innfarn et al. \cite{mesh} sets the precedent for watermarking 3D models using deep learning, but more work is required before a generalised, practically-applicable framework for 3D watermarking can be developed.

Similarly, audio watermarking faces a comparable situation. While there are existing traditional methods for audio watermarking, and GAN-based frameworks for audio generation \cite{audiogan,audiogansurvey}, there is currently a lack of exploration of deep learning frameworks specifically designed for audio watermarking. This suggests that machine learning models have the potential to learn audio embedding techniques and apply them to audio databases. As far as we are aware, there are currently no works that investigate deep learning frameworks for audio watermarking. However, given the growing interest in this field, it is likely that there will be forthcoming contributions.

In addition to image and audio watermarking, video watermarking is another promising direction for research. There are existing traditional algorithms for video watermarking \cite{videosurvey}, and deep learning techniques are also being considered, for example in \cite{vidattention}. This paper introduces RIVAGAN, a new architecture for robust video watermarking. The attention-based architecture is robust against common video processing attacks such as scaling, cropping, and compression. In the framework, a 32-bit watermark is embedded into a sequence of frames, and the watermark can be extracted from any individual or collection of frames. The framework uses an attention mechanism to identify robust areas for embedding, and produces watermarked footage that is nearly indistinguishable from human observers (approximately 52\% detection accuracy). There is also DVMark \cite{dvmark} from Google Research, which employs a multiscale design where the watermark is distributed across multiple spatial-temporal scales. It was found to be more robust than traditional video watermarking algorithms (3D-DWT) and the deep learning-based image watermarking framework from HiDDeN \cite{hidden} against video distortions, while retaining a high level of quality. A 3D-CNN that simulates video compression attacks was used during training to achieve high levels of robustness. The framework in DVMark \cite{dvmark} also includes a watermark detector that can analyse a long video and locate multiple short instances of copyrighted content. This could be useful for reliably identifying copyright infringement on online video platforms such as YouTube. 

\subsection{Text Watermarking for Combating Misinformation}

Another promising application for deep watermarking is in watermarking text. As natural language generation technology improves, machine learning models are capable of generating highly fluent language that can fool human detectors \cite{review}. There is growing concern that such models will be used to spread misinformation and `fake news' online. The Adversarial Watermarking Transformer (AWT) developed by Sahar et al. \cite{textwm} is the first end-to-end deep learning model for language watermarking. Language watermarking is inherently more complex than image, video, and audio watermarking because the language itself must be altered, which can cause drastic syntactic and semantic changes. Previous techniques include synonym replacement and altering sentence structure, which rely on fixed rule-based techniques. The aim of such techniques is to achieve high effectiveness, secrecy, and robustness, while sustaining only subtle changes to the text, preserving correct structure and grammar, as well as language statistics. The deep learning model AWT \cite{textwm} uses an attention mechanism and adversarial training to achieve robustness against attacks such as denoising, random changes, and re-watermarking. The model undergoes human evaluation and achieves better results than the state-of-the-art synonym substitution baseline. This technique only works effectively for long pieces of text such as news articles, whereas shorter pieces would require longer relative watermarks, thereby noticeably degrading the original text. For practical scenarios, it is suggested to combine this AWT technique with automated or human fact-checking to reduce the likelihood of false positives. Further research into deep learning-based models for language watermarking is highly important as text-generating models continue to improve and become widely available to potentially malicious actors. This will help to identify misinformation and differentiate generated from genuine text. 

\subsection{Mitigating Watermark Removal Attacks}

As technology for watermark embedding improves, so too does technology for attacking and removing those watermarks. Thus, the process can be regarded as an ever-evolving, adversarial game between content owners and attackers. Currently, deep learning-based techniques for watermark removal exist that are able to remove information robustly embedded using the top frequency domain algorithms \cite{wmremoval}. This technique uses a simple CNN to perform denoising removal attacks, and is able to not only remove the watermark, but also recover the original cover images without significant quality degradation. It was tested on a dataset of watermarked images created using state-of-the-art DCT, DWT, and DFT traditional algorithms in a black-box setting. Although this technique is focused on traditional algorithms, as deep learning techniques for digital watermarking evolve, it is inevitable that adversarial techniques will continue to be developed that aim to remove these watermarks. Therefore, it is important to continue to improve the robustness of watermarking techniques so they can resist these emerging, deep learning-based removal methods. 

Current deep watermarking strategies have been tested against a range of attacks including cropping, pixel dropout, compression, and blurring. However, in most models these attacks are encapsulated by a differentiable attack layer, as was implemented in \cite{hidden,redmark,romark}, meaning that they must support backpropagation, which is not representative of many real-world attack scenarios. However, it should be noted that these models can still simulate JPEG compression, which is non-differentiable, by using differentiable approximations. Promising techniques for improving the scope of attacks used during training include generating attacks using adversarial examples from a trained CNN \cite{DA}, and the use of black-box noise \cite{2stage}, which are from algorithms encapsulated in the image processing software that is difficult to simulate. To achieve optimal robustness, it is important to train models on attacks generated from an adversarial network to improve results, rather than generating attacks from a fixed pool of differentiable attacks, as was done in earlier implementations.

\subsection{Watermarking for Protecting Machine Learning Models}

Another important application for digital watermarking is protecting machine learning models as intellectual property. Although the survey has discussed watermarking digital media such as images, audio, and video, machine learning models themselves require increasingly large amounts of computational resources and private training data to train and operate. Therefore, there is a growing need to protect machine learning models as intellectual property. Digital watermarking is just one of many tasks that were once done using traditional algorithms, but are now being offloaded to the effectiveness of machine learning strategies. As this happens, it is important to protect these models from theft and misuse, not only because of the resource expenditure by the owners in creating these models, but because their immense computational capabilities could be used for malicious activities. There are many techniques currently being researched for watermarking machine learning models, most of which rely on embedded identifying information into training datasets \cite{mldw}. However, as was learned through the concept of adversarial examples, even small perturbations in training instances can cause extreme degradation in the model's performance. Therefore, many watermarking strategies also sacrifice the model's classification accuracy. 

A famous example is DeepSigns \cite{deepsigns}, an end-to-end IP protection framework that inserts digital watermarks into deep learning models by embedding the watermark into the probability density function of the activation sets in different layers of the network. It is robust against a range of attacks, including model compression, fine-tuning, and watermark overwriting, which proved challenging for previous model watermarking techniques.

One recent and promising technique for watermarking training datasets is Entangled Watermarking Embeddings \cite{ewe}. Instead of only learning features sampled from the task distribution, the defending model also samples data that encode watermarks when classifying data. Therefore, watermark data is entangled with legitimate training data, so an adversary attempting to remove the watermarks cannot do this without damaging the training data itself, and thereby sacrificing performance. The method uses Soft Nearest Neighbour Loss to increase entanglement, a new loss function. And the method is evaluated in the image and audio domains, showing that with this method an owner can claim with 95\% confidence that model extraction has taken place to produce a similarly performing model by extracting prediction vectors. This technique notably shows robustness against adaptive adversaries, meaning the adversary has knowledge of the watermarking technique being used.

As machine learning models become more ubiquitous across a range of industries, it is important to implement digital watermarking strategies within the models themselves so that the owner's private information remains secure. However, there will inevitably be technology developed to remove watermarks, even from machine learning models. For example, REFIT is a recent unified watermark removal framework based on fine-tuning. It does not require knowledge of the watermarks being removed, and is effective against a wide range of current watermarking techniques \cite{refit}.

\subsection{Watermarking for Launching Backdoor Attacks}

Deep neural networks have been proven vulnerable to backdoor attacks, where triggers can be embedded into DNNs through data hiding and they can trick the model into producing unexpected behaviour with crafted triggers. For instance, watermarks can be embedded into the training examples of machine learning models in order to cause inaccurate classifications when the model is deployed. Many third-party cloud computing providers, such as Google, Microsoft, Azure, and Amazon, provide Machine Learning as a Service (MLaaS) to train machine learning models. A malicious party can embed watermarks into training data images and train the model to misclassify such examples, either randomly (random target attack), or mislabel as a different example (single target attack). As more third-party providers offer MLaaS, and machine learning models become more ubiquitous, backdoor attacks on neural networks are certain to become more common. Therefore, it is important to be able to detect triggers embedded in training examples. As digital watermarking technologies become more advanced through deep learning as discussed in this survey, this will become more difficult. In this case, the improving watermark removal techniques as discussed could be pursued for a benign rather than malicious purpose. Although existing works discuss visible watermarks embedded in training images \cite{badnets, trojan}, there are growing efforts to construct invisible backdoor attacks that cannot be detected by a human moderator, or by automated backdoor detection techniques \cite{backdoor, li2021hidden}. Although the techniques presented in this survey are promising for digital IP protection purposes, they will inevitably be utilised for malicious purposes such as embedding undetectable, invisible backdoors in machine learning models. 

\subsection{Deepfake Detection and Identification}

Synthetic media technologies are rapidly advancing, and it is more painless to generate media such as images, audios, and videos that look and sound increasingly realistic. Since deepfakes often present a person saying or doing something they have not done or said, people need to be able to identify the original source of the media that has been manipulated. Similarly, it is important to be able to identify a piece of media as synthetic in the first place, without malicious parties removing this identifying tag and presenting the synthetic media as genuine. To this end, the identification and detection of synthetic media become a promising application for digital watermarking. 

A recent paper presents DeepTag, an end-to-end deep watermarking framework that includes a GAN simulator that applies common distortions to facial images \cite{deeptag}. The watermark can be recovered to identify the original unaltered facial image. In future, as regulations surrounding deepfakes arise, watermarking techniques that are robust to GAN-based distortions will become increasingly important. A connected application is embedding watermarks into synthetic media so that they can be easily identified as such. 

\subsection{Malware using Steganography}
Digital steganography can be used maliciously to spread malware to victim technology. The ability to hide an executable file within an image or audio file gives attackers an easy attack vector to target unaware users. There have also been known cases of attackers using steganography to pass data through unsuspecting platforms. They can easily set a time for the receiver and either upload or update an image temporarily. At this set time the receiver can download and save the photo, decode the message, and then the attacker can restore the image to the original or delete the image. This could be almost impossible to detect when third parties are unaware of where the attack will take place, and would be even more unlikely to catch the act if the stego-image used was highly imperceptible. These high levels of imperceptibility can now be achieved easily through modern deep learning approaches. 

Steganalysis could be implemented into antivirus software that not only scans images but scans sites before entering. Though this would be a difficult task due to the large amount of power it would require, constantly scanning almost every website or item on the web page just \textit{in case} there is malware present. There could be a filter that decides when steganalysis could be used such as situations when users decide to continue onto an already scanned suspicious website. Currently, the possibilities are limited, but the implementation of steganalysis in antivirus software may be essential in the future as steganography techniques both improve in performance and become more widely accessible to the public.

\section{Conclusion}
\label{section:9}

This survey has provided an extensive overview of current deep learning techniques for data hiding, encompassing watermarking and steganography methods. Through analysis of network architecture and model performance, the survey has demonstrated how digital watermarking and steganography share a common goal of embedding information in digital media, and how both can benefit from deep learning techniques. Additionally, the survey explored future research directions and highlighted the potential for this field to revolutionize the protection of digital IP and communication security in Responsible AI software industries. As deep learning techniques continue to advance, they are expected to surpass traditional algorithms in all types of media, ultimately enhancing the accountability and safety of AI. This promising field holds great potential and is expected to have a significant impact on digital security.

\section*{Acknowledgments}

We would like to thank Wendy La for the discussion and feedback on the paper.

\ifCLASSOPTIONcaptionsoff
  \newpage
\fi

\bibliographystyle{IEEEtran}
\bibliography{IEEEabrv,Bibliography}

\vfill

\end{document}